\documentclass[10pt,twocolumn,letterpaper]{article}

\usepackage[pagenumbers]{cvpr} 

\usepackage{microtype}
\usepackage[accsupp]{axessibility}

\usepackage{multirow}
\usepackage{pifont}
\usepackage[percent]{overpic}

\newcommand{\cmark}{\ding{51}}
\newcommand{\xmark}{\ding{55}}

\definecolor{light-gray}{gray}{0.5}
\definecolor{pretty-blue}{RGB}{0, 113, 188}
\def\ie{\emph{i.e.}}
\definecolor{linecolor}{gray}{.895} 
\definecolor{tabhead}{RGB}{213,232,212}
\definecolor{SupMet}{RGB}{218,232,252}

\newcommand{\relativeimpZ}[1]{\small{0.0}}

\definecolor{lightblue}{RGB}{167,224,227}
\definecolor{lightred}{RGB}{248,206,204}
\definecolor{lightgreen}{RGB}{213,232,212}
\definecolor{darkgreen}{RGB}{90,133,95}

\usepackage{todonotes}

\usepackage{colortbl}

\newcommand{\ourmethod}{Fase3D\xspace}

\def\supp{\emph{Supp. Mat.}}

\definecolor{tokenizer}{HTML}{FAC7DA}
\definecolor{FFTenhancer}{HTML}{D9F2D0}
\definecolor{graph_merging}{HTML}{C1E5F5}
\definecolor{llm}{HTML}{DBC4F0}
\definecolor{gfm}{HTML}{FBE3D6}
\definecolor{mygreen}{RGB}{118,199,86}
\newcommand{\gain}[1]{\textcolor{mygreen}{+#1}}

\definecolor{cvprblue}{rgb}{0.21,0.49,0.74}
\usepackage[pagebackref,breaklinks,colorlinks,allcolors=cvprblue]{hyperref}

\def\paperID{***} 
\def\confName{CVPR}
\def\confYear{2026}

\title{Efficient Encoder-Free Fourier-based 3D Large Multimodal Model}

\author{Guofeng Mei$^{1}$ \quad Wei Lin$^2$ \quad Luigi Riz$^1$ \quad Yujiao Wu$^3$ \quad Yiming Wang$^1$ \quad Fabio Poiesi$^1$\\
$^1$Fondazione Bruno Kessler, Italy \quad
$^2$JKU Linz, Austria \quad
$^3$CSIRO, Australia \\
{\tt\small \{gmei,luriz,ywang,poiesi\}@fbk.eu,\tt\small wlin2021at@gmail.com,\tt\small yujiao.wu@csiro.au}
}

\begin{document}
\maketitle
\begin{abstract}
Large Multimodal Models (LMMs) that process 3D data typically rely on heavy, pre-trained visual encoders to extract geometric features. While recent 2D LMMs have begun to eliminate such encoders for efficiency and scalability, extending this paradigm to 3D remains challenging due to the unordered and large-scale nature of point clouds. This leaves a critical unanswered question: How can we design an LMM that tokenizes unordered 3D data effectively and efficiently without a cumbersome encoder?
We propose \ourmethod, the first efficient encoder-free \underline{F}ourier-b\underline{ase}d \underline{3D} scene LMM. \ourmethod tackles the challenges of scalability and permutation invariance with a novel tokenizer that combines point cloud serialization and the Fast Fourier Transform (FFT) to approximate self-attention. This design enables an effective and computationally minimal architecture, built upon three key innovations: First, we represent large scenes compactly via structured superpoints. Second, our space-filling curve serialization followed by an FFT enables efficient global context modeling and graph-based token merging. Lastly, our Fourier-augmented LoRA adapters inject global frequency-aware interactions into LLM backbones at a negligible cost. \ourmethod achieves performance comparable to encoder-based 3D LMMs while being significantly more efficient in computation and parameters. Project website: \url{https://tev-fbk.github.io/Fase3D}.
\end{abstract}    
\section{Introduction}

\begin{figure}[t]
    \centering
    \includegraphics[width=\linewidth]{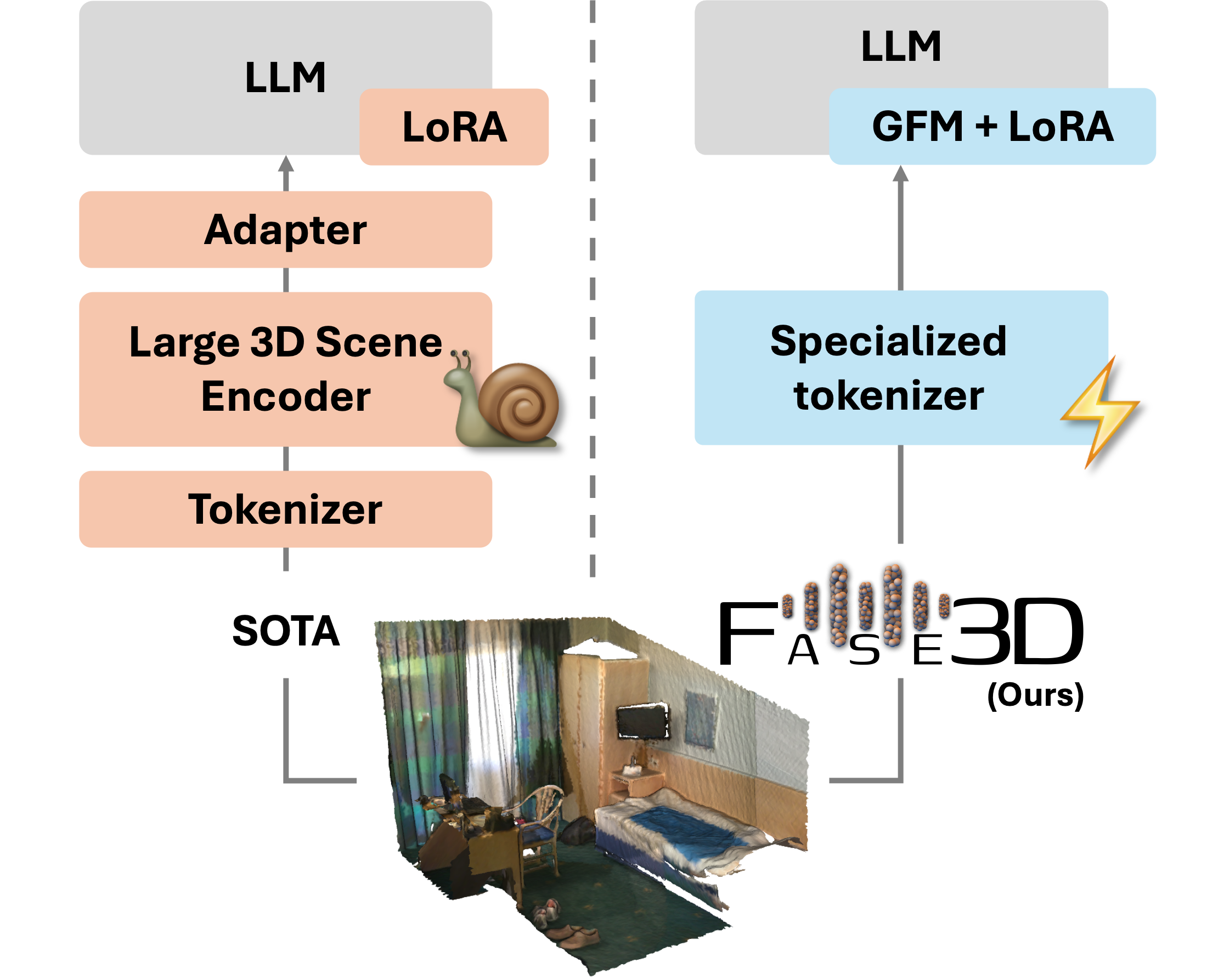}
    \vspace{-7mm}
    \caption{\ourmethod's contribution overview. 
    Mainstream 3D~LMMs are based on computationally-heavy scene encoders to extract geometric features before alignment with the LLM. In contrast, our method (\ourmethod) employs a lightweight Fourier-based tokenizer to process raw point clouds directly and introduces Fourier-augmented LoRA adapters, which infuse global frequency-aware context into the LLM without additional computational overhead.
    }
    \label{fig:teaser}
    \vspace{-4mm}
\end{figure}

\label{sec:intro}

A typical practice in 3D Large Multimodal Models (LMMs) involves using pre-trained vision encoders (\eg, CLIP~\cite{clip} or Sparse 3D U-Net~\cite{deng20253d}) to extract high-level visual semantics, which are then mapped into the language model's embedding space~\cite{li2024llava, mei2025perla}. 
These encoders are effective, but impose substantial computational overhead and limit input flexibility. 
To improve scalability and efficiency, recent works on 2D LMMs have explored vision encoder-free architectures, such as EVE/EVEv2~\cite{diao2024unveiling, diao2025EVEv2} and Mono-InternVL~\cite{luo2025mono}.
Although constructing these models is challenging due to the lack of large-scale vision pre-training, specialized modules, such as visual experts~\cite{luo2025mono} or modality-aware components~\cite{diao2025EVEv2}, have enabled encoder-free designs to approach the performance of encoder-based counterparts.
In the 3D domain, however, encoder-free scene LMMs remain largely unexplored. Using multi-view 2D workarounds is insufficient, as sacrificing the illumination invariance of 3D sensors degrades performance in low-light environments~\cite{zhu2024llava3d}. Moreover, directly transferring 2D encoder-free architectures to native 3D data is non-trivial. Unlike regular pixel grids, point clouds are unordered and irregular, requiring specialized permutation-invariant operators~\cite{qi2017pointnet++,li2025cross,mei2024unsupervised} or serialization techniques~\cite{wu2024point,mei2024vocabulary}. This gap raises a critical question: \textit{how can we design encoder-free 3D LMMs that operate directly on point clouds while remaining both effective and efficient?}
Designing such models demands addressing unique 3D challenges. Since vision encoder-free architectures lack traditional pre-training, they must incorporate explicit inductive bias to robustly tokenize the inherent unordered nature of point clouds. This specialized tokenizer must have minimal learnable parameters for efficiency. Furthermore, given the arbitrary length and massive scale of point clouds, the overall LMM architecture must be computationally and memory efficient.

In this paper, we introduce an efficient, encoder-free, Fourier-based 3D scene LMM, named \ourmethod, that effectively processes scene-level point clouds. 
\ourmethod introduces a novel interaction by viewing token processing as a synthesis between the spatial and frequency domains. 
This dual-domain aggregation effectively captures global and local semantic and geometric information while inherently reducing complexity by approximating self-attention. 
Its core mechanisms are point cloud serialization~\cite{wu2024point} and the Fast Fourier Transform (FFT), which together yield a highly effective and efficient tokenizer. 
FFT is a powerful operator that can approximate self-attention and aggregate global context while being computationally efficient~\cite{fourier_transformer}. We leverage frequency domain processing across several stages of our pipeline. 
We first apply FFT to precomputed, serialization-based geometry superpoints to generate context-aware candidate tokens.
We then introduce a sparse graph-based token merging formulation that adaptively reduces the token set, significantly lowering GPU cost.
Lastly, we propose a novel strategy for training the LLM by augmenting LoRA layers~\cite{hu2022lora} in the frequency domain of low-pass filtered input tokens. 
We evaluate our method on 3D dense captioning and question answering.
With substantially fewer activated visual parameters, \ourmethod achieves comparable performance to state-of-the-art approaches, like LL3DA~\cite{chen2024ll3da} and PerLA~\cite{mei2025perla}, on ScanQA~\cite{azuma2022scanqa}, SQA3D~\cite{ma2022sqa3d}, ScanRefer~\cite{chen2020scanrefer}, and Nr3D~\cite{achlioptas2020referit3d}.

\noindent To summarize, our main contributions are:
\begin{itemize}
\item We present \emph{\ourmethod}, the first scene-level encoder-free 3D LMM that eliminates dedicated 3D encoders and instead integrates superpoint tokenization, positional encoding, and FFT-based augmentation into a standard LLM.
\item We propose an \emph{FFT context enhancer} that leverages space-filling curves (SFCs) to enable efficient frequency-domain mixing and compact token merging.
\item We introduce an efficient \emph{sparse $k$NN superpoint-graph} construction based on space-filling curve ranking, providing a structured yet lightweight representation.
\item We design a \emph{Fourier-augmented LoRA adapter} that enriches the LLM’s internal layers with global context modeling at negligible computational and parameter cost, preserving the monolithic philosophy.
\end{itemize}
\section{Related Work}
\label{sec:related}

\noindent\textbf{Encoder-based 3D LMMs.}  
Early work on 3D vision and language understanding relies on specialized geometric feature extractors to bridge spatial structure and semantics.
For Visual Question Answering (VQA), ScanQA~\cite{azuma2022scanqa} proposed a baseline by pairing point-cloud features with text via dedicated encoders.
Subsequent methods focused on cross-modal fusion and unified tasks: 3D-LLM~\cite{hong20233dllm} introduced pre-training to strengthen cross-attention among point clouds, images, and text. 
Chatscene~\cite{huang2024chat} embeds segmented 3D objects into LLM-interpretable tokens.
LL3DA~\cite{chen2024ll3da} and PerLA~\cite{mei2025perla} unified captioning, QA, and grounding by coupling a point encoder with Q-Former adapters for alignment. 
To tackle large scene understanding, methods like LSceneLLM~\cite{zhi2025lscenellm} and MICAS~\cite{shao2025micas} were proposed, using techniques such as adaptive region selection, scene magnification, and multi-grained sampling for improved detail capture and efficient grounding. 
Other approaches, such as DAC~\cite{wang2025describe}, offer a simpler CLIP+MLLM recipe for open-set 3D object retrieval. 
In parallel, many works extend the popular 2D LLaVA~\cite{liu2024llava} architecture: 3D-LLaVA~\cite{deng20253d} integrates 3D encoders aligned via instruction tuning for open-vocabulary QA and grounding. 
Similarly, methods like LISA~\cite{li2024lisa} and SceneLLM~\cite{fu2025scene}, LLaVA-3D~\cite{zhu2024llava3d}, and SceneVerse~\cite{jia2024sceneverse} lift 2D priors (like mask proposals) into the 3D domain for holistic scene understanding. 
While these methods establish strong baselines, their dependence on computationally expensive encoders constrains input resolution and scalability. 
Moreover, their resulting feature embeddings often remain semantically misaligned with the reasoning capabilities of LLMs~\cite{tang2025exploring}. 
Most existing systems still require dedicated 3D encoders or projection modules to process geometric information effectively.

\noindent\textbf{Encoder-free LMMs.}
Recent advances in monolithic 2D architectures, which integrate perception and reasoning within a single decoder-only Transformer, have inspired a shift toward encoder-free Large Multimodal Models (LMMs). 
Examples include SOLO~\cite{chen2024solo}, Fuyu-8B~\cite{fuyu-8b}, EVE/EVEv2~\cite{diao2024unveiling,diao2025EVEv2}, and Mono-InternVL~\cite{luo2025mono}. 
These models eliminate modality-specific vision backbones by mapping visual inputs directly into the LLM’s token space through lightweight projections. 
Extending this paradigm to 3D data is non-trivial: point clouds are large, sparse, and unordered, making naïve serialization computationally expensive and permutation-variant, which disrupts instance coherence and global context modeling.
Early explorations demonstrate the feasibility of encoder-light designs for object-level reasoning under instruction tuning. 
ENEL~\cite{tang2025exploring} adopts a lightweight hierarchical tokenization strategy to reduce reliance on heavy vision Transformers in ShapeLLM~\cite{qi2024shapellm} and PointLLM~\cite{xu2024pointllm}.
However, these approaches still struggle to scale to full scenes and capture long-range, cross-instance dependencies. 
Our \ourmethod directly addresses these limitations by making encoder-free 3D modeling practical for scene-level reasoning, while tackling the challenges of token ordering, scalability, and global context integration.

\section{\ourmethod}\label{sec:method}

\begin{figure*}[t]
    \centering
    \includegraphics[width=\linewidth]{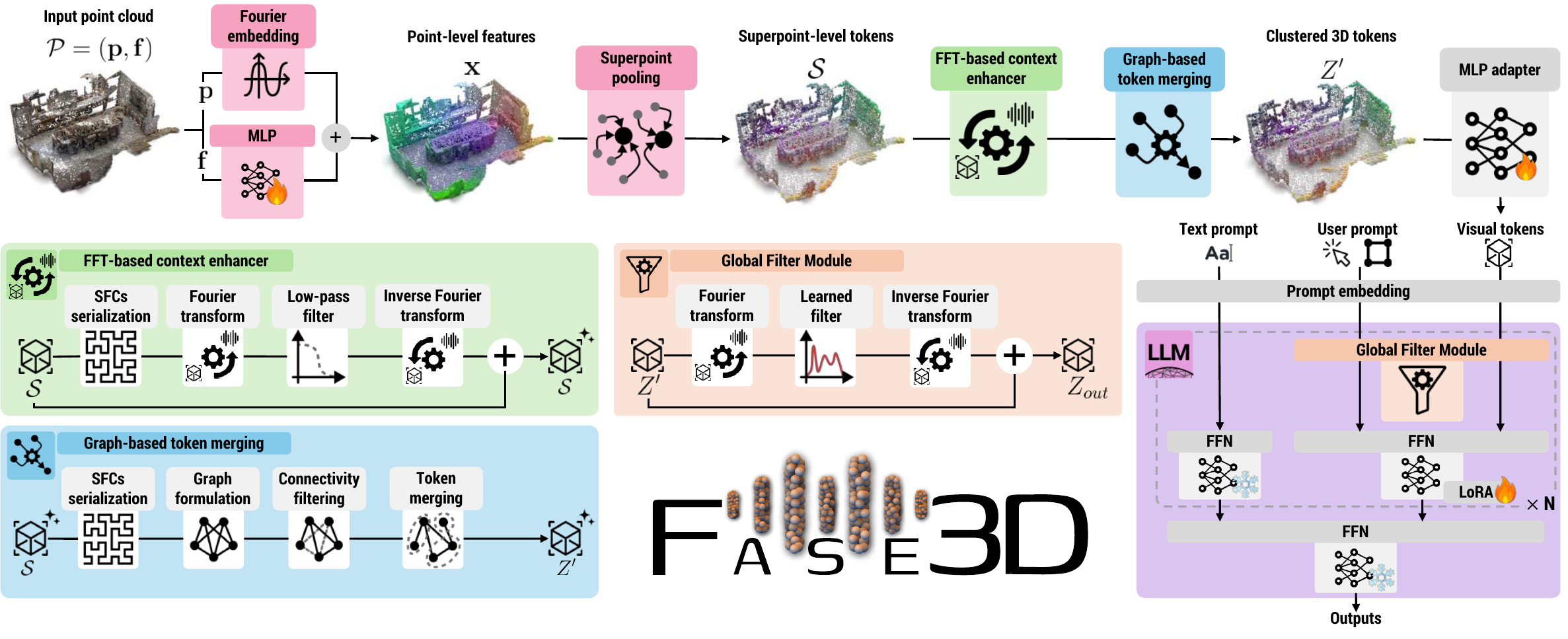}
    \vspace{-7mm}
    \caption{
    The \ourmethod pipeline. A lightweight tokenizer ({\color{tokenizer}{\Large$\bullet$}}) produces $M$ superpoint tokens, which are refined by an FFT-based context enhancer ({\color{FFTenhancer}{\Large$\bullet$}}). A graph is then constructed, and a token-merging block ({\color{graph_merging}{\Large$\bullet$}}) compresses the tokens into $T$ compact 3D tokens ($T < M$). Finally, an LLM ({\color{llm}{\Large$\bullet$}}) with an FFT-based global filter ({\color{gfm}{\Large$\bullet$}}) processes these tokens together with textual and user prompts.
    }
    \label{fig:overview}
    \vspace{-3mm}
\end{figure*}

\ourmethod is a vision encoder-free LMM that relies on a specialized tokenizer to abstract input point clouds into a set of tokens for a decoder-only LLM~\cite{qwen25}. 
\ourmethod is designed to be parameter- and compute-efficient.
It progressively reduces the input token count while enhancing their semantic and spatial information (Fig.~\ref{fig:overview}).
We first compute point-level features with a lightweight multi-layer perceptron~(MLP) that capture only local context and partition the point cloud into $M$ superpoints via geometric clustering~\cite{landrieu2018large} (\S\ref{sec:superpoint_tokenization}). 
We then average-pool the features within each superpoint to obtain candidate 3D tokens.
We serialize the superpoints and employ the Fast Fourier Transform (FFT) on the sequence to further encapsulate contextual information.
This FFT-based context enhancer applies frequency weighting to capture context (\S\ref{sec:fft_enhancer}).
We introduce a graph-based token merging strategy to further aggregate semantic and spatial information into $T$ clusters ($T{<}M$), yielding the final $T$ visual tokens for the 3D scene~(\S\ref{sec:graph_merging}). 
These tokens, together with the text and user prompts, are fed into a prompt embedding~(\S\ref{sec:prompt}) whose outputs serve as inputs to the LLM.
Lastly, we enrich global context by training Fourier-augmented LoRA adapters on the weighted tokens in the frequency domain (\S\ref{sec:lora_fft}).

\subsection{Superpoint-based token initialization}\label{sec:superpoint_tokenization}

To reduce the number of input tokens for the LLM, we employ geometric clustering and produce superpoints~\cite{mei2025perla}.
We use average pooling to aggregate information from neighboring points into a corresponding superpoint embedding~\cite{robert2023efficient, deng20253d}. 
Specifically, let $\mathcal{P} {=} \{(\mathbf{p}_i, \mathbf{f}_i)\}_{i=1}^N$ be the input point cloud, where $\mathbf{p}_i\in\mathbb{R}^{3}$ are the 3D coordinates and $\mathbf{f}_i\in\mathbb{R}^{\mathbf{c}_{in}}$ are additional features (\eg color and normal). 
$\mathcal{P}$ is partitioned into $M$ superpoints $\mathcal{Q}$ centered at $\mathbf{C} {=} \{\mathbf{c}_i\}_{i=1}^M$ via geometric clustering~\cite{landrieu2018large}, where $\mathbf{c}_i\in\mathbb{R}^3$ are the centers.
For each superpoint, we compute a token of dimension $d$. 
Specifically, we first tokenize $\mathcal{P}$ into embeddings $\mathbf{X}^{(0)} \in \mathbb{R}^{N \times d}$. 
For each point $\mathbf{p}_i$, we project $\mathbf{f}_i$ into a $d$-dimensional feature via a shallow learnable multilayer perceptron (MLP), yielding the token $\mathbf{x}^{(0)}_{feat} \in \mathbb{R}^{d}$. 
In parallel, we encode $\mathbf{p}_i$ using a non-parametric Fourier feature embedding of varying frequencies, obtaining $\mathbf{x}^{(0)}_{coor} \in \mathbb{R}^{N \times d}$. 
We obtain the point-level token $\mathbf{x}^{(0)} {=} \mathbf{x}^{(0)}_{feat} + \mathbf{x}^{(0)}_{coor}$.
Lastly, we derive the superpoint-level tokens $\mathbf{S} \in \mathbb{R}^{M \times d}$ by average pooling their associated point-level tokens:
\begin{equation}
\mathbf{S} = \text{SptPool}(\mathbf{X}^{(0)}, \mathcal{Q}) \in \mathbb{R}^{M \times d},
\end{equation}
where \emph{SptPool} stands for superpoint based average pooling.

\subsection{Fourier-based context enhancer}\label{sec:fft_enhancer}

Since the initialized superpoint tokens $\mathbf{S}$ only capture local information, we designed a lightweight token enhancer module that injects global context by operating in the frequency domain \cite{fourier_transformer}. 
Prior ``frequency'' methods typically use either (i) voxel/grid-based 3D FFT, which is costly at scene scale, or (ii) Graph Fourier Transform (\eg, PointGST~\cite{liang2025parameter}) that requires explicit graph construction and Laplacian operations ($O(M^2)$).
In contrast, we first serialize the superpoints~\cite{wu2024point} into a 1D sequence, enabling the application of FFT-based processing. 
The FFT operates with complexity $O(M \log M)$ over $M$ tokens and adaptively reweights spectral components to enable context mixing and capture global layout information (\eg, object groupings).
The inverse FFT (iFFT) then produces a spatially varying global context field that is fused back into the original tokens via residual addition. 
This design enriches each token with both local and global context, promoting long-range reasoning. To mitigate ordering bias from a single sequence, we adopt \emph{multi-curve serialization} with varied axis orderings to diversify 1D adjacencies.

\noindent\textbf{Token serialization.}
We serialize tokens to apply FFT.
Specifically, we map the coordinates $\mathbf{C}$ into a \emph{locality-preserving} 1D sequence using \emph{SFCs}~\cite{wu2024point,sagan2012space}, centering on four representatives (denoted as $\pi=\{\pi_i\}_i$): the z-order curve, transpose z-order curve, Hilbert curve and the transpose Hilbert curve.
We can reorder the superpoint tokens $\mathbf{S}$ into a 1D sequence $\mathbf{S}[\pi_i]$ where 3D locality is preserved.

\noindent\textbf{FFT-based token enhancer.}
We perform spectral mixing over the serialized token sequence using a real-valued frequency transform to aggregate contextual information~\cite{fourier_transformer}. 
We describe the module using FFT/iFFT notation; in implementation, it can be realized with either rFFT-based operations or a DCT-style transform for improved efficiency on real-valued inputs. 
Let $\mathcal{F}$ and $\mathcal{F}^{-1}$ denote the 1D FFT and its inverse (iFFT), respectively, applied along the token axis. 
Additional details are provided in the \supp{}
For each traversal $\pi_i$, we apply a frequency transform to the sorted tokens and modulate informative frequency components by
\begin{equation}\label{eq:fourier_mixing}
\mathbf{S}'(\pi_i)=\mathcal{F}^{-1}\!\left(\mathcal{F}\!\left(\mathbf{S}(\pi_i)\right)\odot \mathbf{G}_v\right),
\end{equation}
where $\mathbf{G}_v$ is a learnable non-negative frequency-domain gate, and $\odot$ denotes element-wise multiplication. 
To obtain position-aware mixing, we apply \eqref{eq:fourier_mixing} to overlapping windows of length $L_w{=}128$ and stride $L_s{=}L_w/2$ on $\mathbf{S}(\pi_i)$, and reconstruct by overlap--add with squared-Hann weights~\cite{pfister2017discrete}. 
This yields localized spectral aggregation while maintaining a complexity of $O(L_w\log L_w)$ per window.
We process all curve traversals, restore the original token order via inverse permutations, and fuse them by uniform averaging:
\begin{equation}
\tilde{\mathbf{S}}
=\frac{1}{|\pi|}\sum_{\pi_i}\mathbf{S}'\left(\pi_i\right).
\end{equation}
We fuse the enhancement with a residual as $\mathbf{S}\leftarrow \mathbf{S}+\tilde{\mathbf{S}}$.

\subsection{Graph-based token merging}\label{sec:graph_merging}
To improve computational efficiency, we reduce the number of tokens by merging superpoints into a compact set of informative tokens that better align with object-level structures. Specifically, we aggregate the initial superpoints $\mathbf{S}$ into $T$ tokens using a lightweight module with only a few learnable parameters, and then feed them into the LLM.
Furthermore, we optionally perform spectral clustering on the superpoint graph to generate 3D masks for dense captioning. This graph-based formulation removes the need for an explicit detection stage, as commonly adopted in existing 3D LMMs~\cite{deng20253d, mei2025perla}, which rely on learned mask proposals (\eg, Mask3D~\cite{schult2023mask3d}). In contrast, our method leverages purely geometry-driven superpoints without any learned mask generation. 
Despite its simplicity, it achieves strong empirical performance.

We model superpoints and their relationships as a graph, where superpoint tokens serve as nodes and their spatial relationships define edges, promoting semantically and spatially coherent representations. 
We construct a sparse superpoint graph $\mathcal{G}=(\mathcal{V},\mathcal{E})$ at the point cloud $\mathcal{P}$ level.
This graph is computed once and used as a \emph{topological prior} for the merging stage. The vertices $\mathcal{V}$ correspond to the $M$ superpoints, while the edges $\mathcal{E}$ encode their geometric relationships.

\noindent\textbf{Neighbor searching via window voting.}
Delaunay triangulation is common for point-cloud graph construction, but can be computationally expensive for large-scale point sets~\cite{robert2023efficient}.
Instead, we connect superpoints via a point-level window-voting scheme along 1D orderings induced by SFCs. Analogous to our superpoint serialization, we serialize all points into four SFC traversals, which avoids explicit radius or $k$-NN queries. For each curve, we scan the sorted index list and sample anchor positions $p$ with stride $s_r$. 
Let $i$ be the point index at the center of window $\mathcal{W}(p)$, and let $j$ be the point index for any other position $q \in \mathcal{W}(p)$.
Let $s_i$ and $s_j$ denote the superpoint indices points $i$ and $j$ are assigned to, respectively. 
If both points belong to valid superpoints and these superpoints are different ($s_i {\neq} s_j$), we cast a vote for the edge $(s_i, s_j)$ as $v_{s_i,s_j} {\leftarrow} v_{s_i,s_j} {+} 1$.
Aggregating these votes across all curves yields a sparse set of superpoint pairs with integer vote counts, which defines the graph adjacency. Graph construction complexity is analyzed in the \supp

\noindent\textbf{Graph-based token merging.}
We introduce a token merging module to further reduce the number of tokens from $M$ to $T$ ($T<M$). The module compresses the superpoint features $\mathbf{S}$ by jointly exploiting the global context provided by the curve-based serialization order $\pi$ and the local connectivity encoded in the superpoint graph $\mathcal{G}$, yielding a compact token set that preserves both long-range context and local geometric structure. To this end,  we adopt a lightweight point-seeded graph pooling module that aggregates superpoint features over sparse local neighborhoods on the superpoint graph. 
Concretely, we first sample $T$ anchor points $\{\mathbf{a}_t\}_{t=1}^T$ from the input point cloud by farthest point sampling (FPS), and map them to their corresponding superpoints as initial seeds $\{s_t\}_{t=1}^T$. To improve coverage, duplicated or invalid seeds are removed by graph-aware non-maximum suppression, and the suppressed slots are replaced with uncovered valid superpoints. For each seed superpoint $s_t$, we define a sparse local support set as $\mathcal{N}_t=\{s_t\}\cup\mathcal{N}(s_t)$, where $\mathcal{N}(s_t)$ denotes the 1-hop neighbors of $s_t$ on the superpoint graph.
For each superpoint $i\in\mathcal{N}_t$, we use the graph connectivity weight as the pooling prior. Specifically, the edge weight encodes the affinity between neighboring superpoints, computed from their feature similarity and further modulated by the multi-curve voting strength. We define $\tilde{w}_{it}=1$ if $i=s_t$, and $\tilde{w}_{it}=w^{\mathrm{graph}}_{s_t i}$ otherwise, where $w^{\mathrm{graph}}_{s_t i}$ denotes the edge weight between the seed superpoint $s_t$ and its neighbor $i$ on the superpoint graph. The normalized pooling weight is then given by 
$w_{it}=\frac{\tilde{w}_{it}}{\sum_{j\in\mathcal{N}_t}\tilde{w}_{jt}+\epsilon}$, 
where $\epsilon=10^{-5}$ is a small constant to avoid division by zero. Based on these weights, the pooled token feature is computed as $\mathbf{z}_t^{\mathrm{pool}}=\sum_{i\in\mathcal{N}_t} w_{it}\mathbf{s}_i$, where $\mathbf{s}_i$ denotes the feature of superpoint $i$. We keep the anchor positions fixed and only pool local superpoint features, \ie, $\mathbf{c}_t'=\mathbf{a}_t$. In this way, the token position is determined by point-level spatial coverage, while the token content is read out from the local superpoint graph neighborhood. After local normalization and projection, we obtain the final merged token representation $\mathbf{z}_t=\phi(\mathbf{z}_t^{\mathrm{pool}})$, where $\phi(\cdot)$ denotes a lightweight feature transformation.
Finally, since LLM inputs are inherently sequential and rotary positional embeddings depend on token order, we serialize the merged tokens $\mathbf{Z}'=\{\mathbf{z}_t\}_{t=1}^T$ together with their fixed coordinates $\mathbf{C}'=\{\mathbf{c}_t'\}_{t=1}^T$ along a Hilbert curve, yielding a locality-preserving ordered token sequence for the LLM.

\subsection{Prompt embedding}\label{sec:prompt}
In tasks such as dense object captioning or center-object question answering, the language instruction may include explicit coordinates, instances, or bounding boxes. To handle such inputs, we introduce a \emph{3D coordinate token} that enables the model to incorporate spatial cues into its reasoning process. Concretely, we encode the input coordinates (or box/instance centers) together with their $k$-nearest neighbors using a Fourier positional encoding layer~\cite{chen2024ll3da}. The resulting coordinate tokens are concatenated with the 3D patch tokens and text tokens before being fed into the LLM. This design enables coordinate-aware 3D perception and reasoning.


\subsection{Fourier-augmented LoRA adapter for LLM}
\label{sec:lora_fft}

While LoRA provides a parameter-efficient method for adapting the linear feed-forward network (FFN) layers, the representations $\mathbf{Z}'\in \mathcal{R}^{T\times D}$ fed into them are generated by the frozen, pre-trained backbone. These representations are not explicitly optimized for the downstream task, which can limit the potential of the LoRA updates. We propose to enhance these representations by introducing a lightweight \emph{Global Filter Module (GFM)}. 
The goal is to enrich the token features with globally-mixed information, thus providing a more robust and adaptive input to the LoRA-adapted layers.
Our GFM is inspired by the efficiency of Fourier transforms for global mixing~\cite{lee2021fnet,qin2021fcanet}. 
It operates on each token $\mathbf{z} \in \mathcal{R}^D$ within the sequence $\mathbf{Z}'$. 
Specifically, we first project the feature vector into the frequency domain, apply a learnable filter, and then project back to compute the mixed representation as $\mathbf{z}_{\text{mixed}} {=} \text{iFFT}(\text{FFT}(\mathbf{z}) \odot \mathbf{G}_t)$, where $\text{FFT}$ and $\text{iFFT}$ are applied along the channel dimension $D$. 
$\mathbf{G}_t \in \mathcal{R}^D$ is a learnable parameter vector that acts as the filter.
The final enhanced feature $\mathbf{z}_{\text{out}}$ fed into the LoRA-adapted layer is then computed as an averaged residual connection as $\mathbf{z}_{\text{out}} = (\mathbf{z} + \mathbf{z}_{\text{mixed}}) / 2$.
This blending of the original and filtered representations provides a rich, globally-aware input to the FFN while maintaining a stable learning dynamic. 
This module remains highly efficient, introducing only $D$ learnable parameters (the filter $\mathbf{G}_t$). We adopt a multi-head formulation with $N_h$ heads to reduce the effective computational cost, resulting in $O\left(\tfrac{D}{N_h} \log \tfrac{D}{N_h}\right)$ complexity.

\subsection{Training}
We adopt a two-stage training strategy inspired by~\cite{mei2025perla,deng20253d}: 
First, we perform general 3D instruction tuning, then we specialize the model on downstream tasks.

\noindent\textbf{Language modeling loss.}
For 3D scene–text pairs, we optimize caption generation with next-token cross-entropy, following the standard LLM setup:
\begin{equation}
\mathcal{L}_{\mathrm{LM}}
= - \frac{1}{\sum_{t}m_t}\sum_{t} m_t\,
\log p_{\theta}\!\big(w_t \mid w_{<t},\mathbf{Z}'\big),
\label{eq:lm_loss}
\end{equation}
where $p_{\theta}$ is the model’s token distribution, $m_t={1}[t\ge t_0]$ masks out non-caption prefix/prompt tokens, and $t_0$ indexes the first caption token. Padding tokens are ignored via $m_t$.

\noindent\textbf{Datasets.}
During training, we use the ScanNet v2~\cite{dai2017scannet} portion from the 3DLLM dataset~\cite{hong20233dllm}, which provides 1,201 training and 312 validation reconstructed indoor scenes. For language supervision, we combine four established sources: ScanQA~\cite{azuma2022scanqa}  and SQA3D~\cite{ma2022sqa3d} for 3D question answering, and ScanRefer~\cite{chen2020scanrefer} together with Nr3D (ReferIt3D)~\cite{achlioptas2020referit3d} for 3D referring expression comprehension/localization.
These components jointly define our benchmark, covering both QA and dense captioning. We follow the official train/val splits and report results on the validation set unless otherwise specified. Detailed statistics are in the \supp
\section{Experiments}

\noindent\textbf{Implementation details.} 
Following \cite{chen2024ll3da,chen2021scan2cap}, we uniformly sample 50k points per scene.  Point features are pooled into superpoint tokens, followed by clustering into 256 tokens. The number of heads is set to $N_h{=}8$.
We use a frozen Qwen2.5-3B-Instruct~\cite{qwen25} language model in \textit{float16}. 
The LoRA configuration is rank $r{=}768$ with scaling $\alpha{=}768$ for the first 8 layers.
We use AdamW \cite{loshchilov2017fixing} with weight decay 0.1 and cosine decay from $10^{-4}$ to 1$0^{-6}$ over $\sim$100k iterations. 
We train with a batch size of 8 for seven days on up to four NVIDIA A100 64GB GPUs. 
For each task, we fine-tune only the parameters for $\sim$30k iterations.

\noindent\textbf{Metrics.} 
We follow the evaluation protocol~\cite{chen2024ll3da,mei2025perla} to evaluate the quality of output responses.
We use the abbreviations C, B-4, M and R for CIDEr~\cite{vedantam2015CIDEr}, BLEU-4~\cite{papineni2002bleu}, METEOR~\cite{banerjee2005meteor}, and ROUGE-L~\cite{lin2004rouge}, respectively.
We report \#Params, indicating the number of parameters activated for 3D scene tokenization, and FLOP as an efficiency measure.

\renewcommand{\arraystretch}{0.9}
\begin{table*}[t]
\centering
\small
\caption{Question answering results on ScanQA~\cite{azuma2022scanqa} and SQA3D~\cite{ma2022sqa3d}. 
\#Param/FLOP: number of activated parameters and Floating Point Operation count required for the encoding/tokenization stage.
Best result is in \textbf{bold}. Second best result is \underline{underlined}.
}
    \label{tab:scanqa}
    \tabcolsep 9pt
    \vspace{-3mm}
    {%
    \begin{tabular}{lccc|ccccc}
        \toprule
        \multirow{2}{*}{Method} & \multirow{2}{*}{LLMs} & \multirow{2}{*}{\#Params $\downarrow$} & \multirow{2}{*}{FLOP $\downarrow$} & \multicolumn{4}{c}{ScanQA (val)} & \multicolumn{1}{c}{SQA3D (test)}\\
        \cmidrule(lr){5-8} \cmidrule(lr){9-9}
        & & & & R$\uparrow$ & M$\uparrow$ & B-4$\uparrow$ & C$\uparrow$ & EM@1$\uparrow$\\
        \midrule
        \multicolumn{9}{c}{\textit{Encoder-based 3D LMMs with point cloud as inputs}} \\
        \midrule
        LL3DA~\cite{chen2024ll3da} & OPT-1.3B~\cite{zhang2022opt} & 118.87M & 40.21 & 37.31 & 15.88 & 13.53 & 76.79 & -\\
        PerLA~\cite{mei2025perla} & OPT-1.3B~\cite{zhang2022opt} & 119.76M & 163.38 & 39.60 & 17.44 & 14.49 & 78.13 & -\\
        3D-LLaVA~\cite{deng20253d} & Vicuna-1.5-7B~\cite{vicuna2023}  & 58.26M & 37.75 & \underline{43.10} & \underline{18.40}  & \underline{17.10} &  \bf92.60  & \bf54.5\\
        \midrule
        \multicolumn{9}{c}{\textit{Encoder-free 3D LMMs  with point cloud as inputs}} \\
        \midrule
        \ourmethod & Qwen2.5-3B~\cite{qwen25} & \bf10.54M & \bf2.04 & 42.56 & 18.24 & \bf17.12 & 90.11 & 53.9\\
        \ourmethod & Vicuna-1.5-7B~\cite{vicuna2023} & \underline{12.11M} & \underline{2.09} & \bf43.37 & \bf18.61 & 16.87 & \underline{91.74} & \underline{54.3} \\
        \bottomrule
    \end{tabular}
    }
    \vspace{-3mm}
\end{table*}

\subsection{Evaluation on 3D question answering (3DQA)} 
3DQA involves answering free-form questions about a 3D environment, requiring the model to reason about objects, attributes, and relationships within the scene. 
We benchmark \ourmethod on ScanQA~\cite{azuma2022scanqa} and also report competitive performance on SQA3D~\cite{ma2022sqa3d}. 
Built on ScanNet, ScanQA contains $\sim$41.4k questions across 800 scenes that probe object recognition and 3D reasoning. 
SQA3D extends this setting with $\sim$20.4k situation descriptions covering 6.8k unique situations from 650 scenes, together with $\sim$33.4k associated questions, placing stronger emphasis on situated and embodied scene understanding. 
We also report results with Vicuna-1.5-7B~\cite{vicuna2023} for a fair comparison with 3D-LLaVA. 
\cref{tab:scanqa} summarizes the results on ScanQA (val) and SQA3D (test). 
\ourmethod achieves performance comparable to 3D-LLaVA on both datasets, while significantly outperforming the other encoder-based baselines. 
Notably, these results are obtained with substantially fewer vision parameters (\#Param: 10.54M--12.11M vs.~58.26M for 3D-LLaVA and $\sim$119M for LL3DA/PerLA) and a much lower FLOP count ($\sim$2.0 vs.~37.75 for 3D-LLaVA, 40.21 for LL3DA, and 163.38 for PerLA).
\begin{figure*}[t]
    \includegraphics[width=\linewidth]{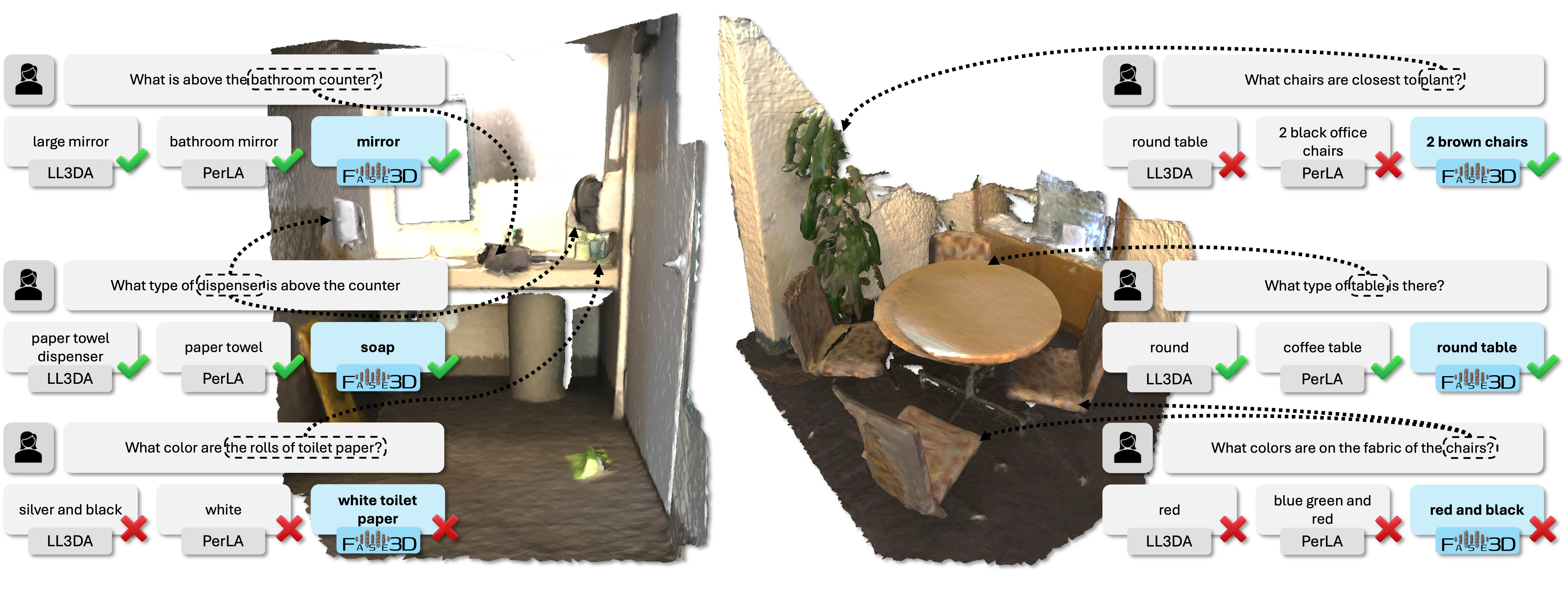}
    \vspace{-9mm}
    \caption{Qualitative results and comparisons between \ourmethod, PerLA~\cite{mei2025perla}, and LL3DA~\cite{chen2024ll3da} on the ScanQA~\cite{azuma2022scanqa} dataset.}
    \label{fig:qualitatives}
    \vspace{-2mm}
\end{figure*}
\cref{fig:qualitatives} compares LL3DA~\cite{chen2024ll3da}, PerLA~\cite{mei2025perla}, and \ourmethod.
In the bathroom scene (left), all methods correctly answer ``\emph{What is above the bathroom counter?}" (mirror). 
For ``\emph{What type of dispenser is above the counter?}", \ourmethod focuses on the smaller \emph{soap} dispenser, while baselines answer \textit{towel}; all are plausible, though ours is more semantically specific.
Yet, all fail on the fine-grained color of the toilet paper rolls, likely due to low texture fidelity and illumination.
In the dining scene (right), \ourmethod identifies attributes and relationships: it answers ``\emph{2 brown chairs}" to ``\emph{What chairs are closest to the plant?}", correctly identifies the \emph{round table}, and recovers the chair-fabric colors ``\emph{red and black}", whereas the baselines miss at least one of these.



\renewcommand{\arraystretch}{0.9}
\begin{table*}[t]
\small
\centering
\caption{Dense captioning results on ScanRefer and Nr3D. 
\#Param/FLOP: number of activated parameters and Floating Point Operation count required for the encoding/tokenization stage. 
Best result in \textbf{bold}. Second best result is \underline{underlined}.
}
\label{tab:3D_dense_captioning}
\vspace{-3mm}
\tabcolsep 3pt
{%
\begin{tabular}{lccc| cccc | cccc}
\toprule
\multirow{2}{*}{Method} & \multirow{2}{*}{Segmenter} & \multirow{2}{*}{\#Params $\downarrow$} & \multirow{2}{*}{FLOP $\downarrow$} & \multicolumn{4}{c|}{ScanRefer} & \multicolumn{4}{c}{Nr3D} \\
\cmidrule(lr){5-8} \cmidrule(lr){8-12} 
& & & & C@0.5$\uparrow$ & B-4@0.5$\uparrow$ & M@0.5$\uparrow$ & R@0.5$\uparrow$ 
& C@0.5$\uparrow$ & B-4@0.5$\uparrow$ & M@0.5$\uparrow$ & R@0.5$\uparrow$ \\
\midrule
\multicolumn{3}{c}{\textit{Encoder-based 3D LMMs}} \\
\midrule
LL3DA~\cite{chen2024ll3da} & \cmark & 118.87M &  80.43 & 65.19 & 36.79 & 25.97 & 55.06 & 51.18 & 28.75 & 25.91 & 56.61\\
PerLA~\cite{mei2025perla} & \cmark & 119.76M & 326.76 & 69.41 & 38.02 & \bf29.07 &  56.80 & \bf55.06 & \bf31.24 & \bf28.52 & \bf59.13 \\
3D-LLaVA~\cite{deng20253d} & \cmark & \underline{58.26M} & \underline{37.75} & \bf78.80 & \underline{36.90} & 27.10 & \bf57.70 & - & - & - & -\\
\midrule
\multicolumn{3}{c}{\textit{Encoder-free 3D LMMs}} \\
\midrule
\ourmethod & \cmark & \multirow{2}{*}{\bf10.54M} & \multirow{2}{*}{\bf2.04} & \underline{78.14} & \bf41.34 & \underline{27.92} & \underline{57.63} & \underline{54.91} & \underline{30.24} & \underline{26.48} & \underline{57.14} \\
\ourmethod  & \xmark &  &  & 70.72 & 37.83 & 26.81 & 56.37 & 52.89 & 29.31 & 26.14 & 56.41 \\
\bottomrule
\end{tabular}
}
\vspace{-3mm}
\end{table*}

\subsection{Evaluation on 3D dense captioning}

3D dense captioning is object-centric and conditioned on regions.
The model localizes object instances and generates fine-grained, attribute-rich descriptions grounded at 3D coordinates. 
To evaluate \ourmethod, we use two proposal variants: with external segmenter Mask3D~\cite{schult2023mask3d} and without an external segmenter, using only our graph-based token-clustering proposals. 
For the variant with our graph-based token merging, we apply spectral clustering on the constructed superpoint graph $\mathcal{G}$ to obtain 48 clusters, which are then treated as proposal instances for evaluation.
We condition the generator with tokens derived from each proposal’s 3D center, and evaluate on ScanRefer~\cite{chen2020scanrefer} and Nr3D~\cite{achlioptas2020referit3d}.
Following prior work~\cite{hong20233dllm,chen2024ll3da}, we report $m@k$IoU, where $m {\in} \{\text{C, B-4, M, R}\}$ and $k$ is the IoU threshold. 
For fair comparison, we list models trained with standard per-word cross-entropy and without extra 3D scene pretraining. 
\cref{tab:3D_dense_captioning} shows that \ourmethod achieves results comparable to 3D-LLaVA in the same setup with the external segmenter  (\textit{Mask3D}) on ScanRefer.
In the \textit{spectral clustering} setting, \ourmethod slightly underperforms 3D-LLaVa, but still maintains similar performance to PerLA. 
On Nr3D, \ourmethod maintains comparable performance to PerLA with the \textit{Mask3D} and \textit{spectral clustering} variants.

\begin{figure*}[t]
    \includegraphics[width=\linewidth]{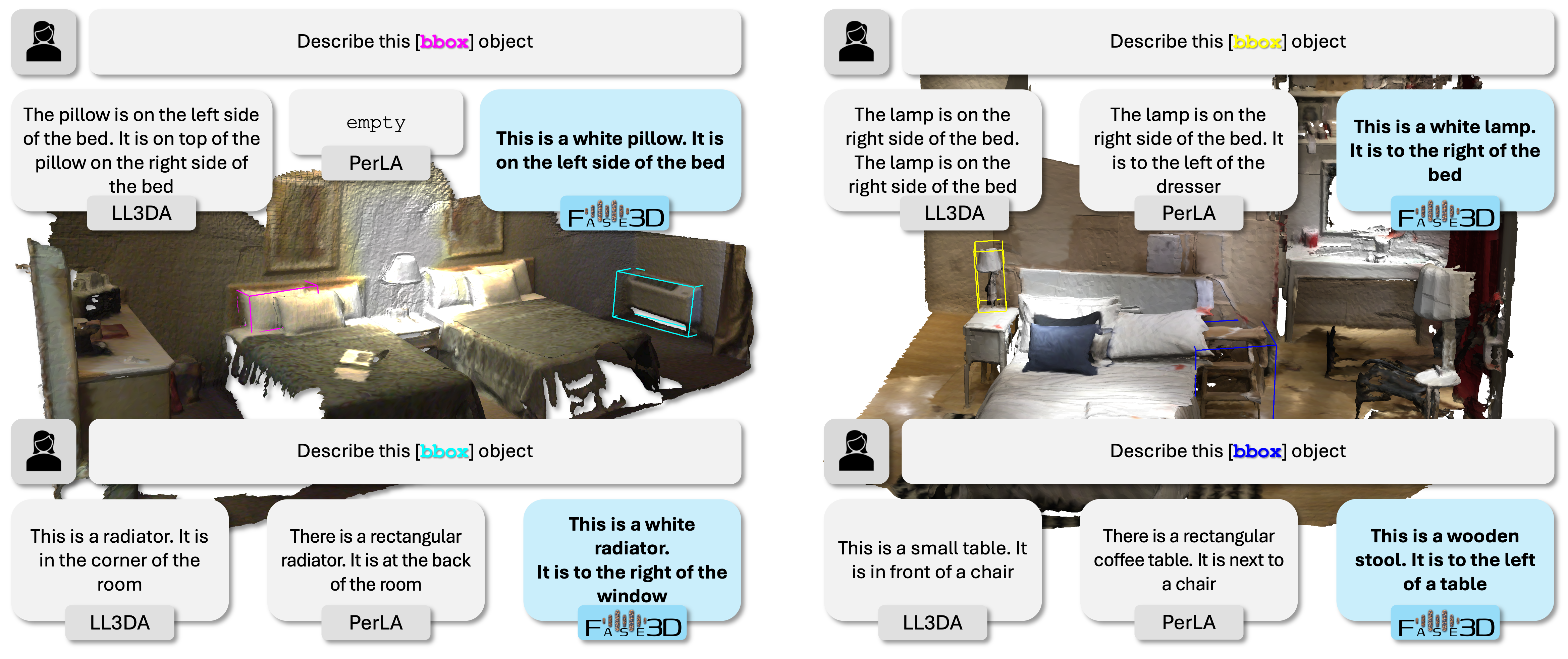}
    \vspace{-7mm}
    \caption{Qualitative comparison between \ourmethod, PerLA~\cite{mei2025perla}, and LL3DA~\cite{chen2024ll3da} on the ScanRefer~\cite{chen2020scanrefer} dataset.}
    \label{fig:qualitatives_2}
    \vspace{-3mm}
\end{figure*}

\cref{fig:qualitatives_2} compares LL3DA, PerLA and \ourmethod on ScanRefer~\cite{chen2020scanrefer}.
In the bedroom scene on the left, 
\ourmethod delivers the most accurate caption for the object in the magenta bounding box, correctly identifying it as a \emph{pillow} positioned on the ``\emph{left side of the bed}''. LL3DA produces a confused description mixing \emph{left} and \emph{right}, and PerLA fails to generate any output.
In the same scan, 
all methods correctly recognize the object within the cyan bounding box as a \emph{radiator}, but only \ourmethod accurately describes its color. PerLA emphasizes its shape, while LL3DA omits descriptive details.
In the second scene (right), all approaches correctly identify the \emph{lamp} in the yellow bounding box despite its small size but misinterpret its spatial relation to the bed. For the blue bounding box, \ourmethod achieves higher semantic accuracy by identifying the object as a \emph{wooden stool}, while LL3DA and PerLA describe it less precisely as a \emph{small table} and a \emph{rectangular coffee table}, respectively.

\subsection{Ablation Studies}\label{sec:ablation}

We assess: 
(i) \emph{patch embedding} alternatives (raw point tokens vs.\ superpoint pooling with or without an FFT-based context enhancer), 
(ii) \emph{LoRA with Fourier residuals} design, and
(iii) \emph{different LLM} backbones. 
Unless otherwise specified, all models are trained from scratch on ScanQA, under identical optimization and data settings. 
We report validation results and vary a single factor at a time, keeping all other components fixed to the default configuration.
See supplementary material for additional ablation studies.

\noindent\textbf{Patch embedding choices.}
We examine how 3D inputs are embedded into tokens for the language head (Qwen2.5-3B). Our pipeline first projects points through a lightweight MLP to point features, then aggregates them via superpoint pooling into superpoint tokens. \cref{tab:ab_patch} ablates three factors: using downsampled raw point tokens (\emph{Point}), adding superpoint pooling (\emph{Superpoint}), and adding our lightweight FFT-based context enhancer. Superpoint pooling shortens the token sequence by roughly one order of magnitude while improving semantic coherence, yielding \gain{3.66} CIDEr over point-only tokens (76.04~$\rightarrow$~79.70). The FFT-based enhancer alone provides \gain{6.93} CIDEr (76.04~$\rightarrow$~82.97). Combining both delivers the strongest ablation result (\gain{10.87} CIDEr; 86.91 total). Training with only raw point tokens is also slower and less stable due to quadratic self-attention. The bottom row reports our \emph{full model with additional pretraining}, further lifting all metrics on the ScanQA val set.
\begin{table}[t]
    \centering
    \small
    \caption{Ablation study of vision embedding modules. Point (downsampled raw point tokens), Superpoint (superpoint pooling), FFT (lightweight FFT-based context enhancer). 
    }
    \label{tab:ab_patch}
    \vspace{-3mm}
    \tabcolsep 7pt
    \resizebox{\columnwidth}{!}
    {%
    \begin{tabular}{ccc|cccc} 
        \toprule
        \multicolumn{3}{c}{Module} & \multicolumn{4}{c}{ScanQA (Validation)} \\
        \midrule
        Point & Superpoint & FFT & R$\uparrow$ & M$\uparrow$ & B-4$\uparrow$ & C$\uparrow$\\ 
        \midrule
        \cmark &            &            & 37.03 & 15.43 & 13.14 & 76.04 \\ 
        \cmark & \cmark &            & 37.18 & 16.38 & 13.96 & 79.70 \\ 
        \cmark &            & \cmark & 39.56 & 17.03 & 15.11 & 82.97 \\ 
        \cmark & \cmark & \cmark & 41.64 & 17.80 & 16.70 & 86.91 \\ 
        \midrule
        \multicolumn{3}{c|}{\textit{Full model with pretraining}} & \textit{42.56} & \textit{18.24} & \textit{18.02} & \textit{90.11} \\
        \bottomrule
    \end{tabular}
    }
    \vspace{-5mm}
\end{table}


\noindent\textbf{LoRA with Fourier residual.}
We compare (i) single-branch LoRA (vision-only or text-only), (ii) shared LoRA on both branches (sLoRA), (iii) decoupled LoRA per branch (dLoRA), and (iv) dLoRA augmented with a Fourier residual branch (+FFT). On ScanQA (val) with Qwen2.5-3B~\cite{qwen25}, adding a Fourier residual to the vision branch yields the best parameter-efficient results: dLoRA+FFT (vision) improves over dLoRA by \gain{4.38} CIDEr, \gain{1.61} BLEU-4, \gain{0.57} METEOR, and \gain{1.68} ROUGE-L (\cref{tab:ab_lora}). However, applying the Fourier residual to both branches reduces the gains on four metrics differently, suggesting frequency-domain cues are most beneficial on the visual pathway. Full end-to-end fine-tuning achieves the highest absolute numbers, but at substantially greater compute and memory; dLoRA+FFT approaches that performance with a fraction of trainable parameters, verifying the effectiveness of our design.

\begin{table}[t]
    \centering
    \small
    \caption{Ablation study of LoRA placement and Fourier residual on ScanQA (Validation)~\cite{azuma2022scanqa}.
        sLoRA: the branches share the \emph{same} LoRA modules.
        dLoRA: the branches use \emph{separate} (decoupled) LoRA modules.
        ``+FFT” adds a parallel Fourier residual branch.
    }
    \label{tab:ab_lora}
    \vspace{-3mm}
    \tabcolsep 4.5pt
    \begin{tabular}{cc|cccc} 
        \toprule
        \multicolumn{2}{c}{Module} & \multicolumn{4}{c}{ScanQA (Validation)} \\
        \midrule
        vision & text & R$\uparrow$ & M$\uparrow$ & B-4$\uparrow$ & C$\uparrow$\\ 
        \midrule
        LoRA  &   -            & 36.98 & 15.37 & 13.35 & 76.45 \\
        -     &   LoRA         & 36.49 & 15.40 & 13.19 & 76.21 \\
        sLoRA & sLoRA          & 37.34 & 16.04 & 12.83 & 78.24 \\
        dLoRA & dLoRA          & 39.96 & 17.23 & 15.09 & 82.53 \\
        dLoRA+FFT & dLoRA      & 41.64 & 17.80 & 16.70 & 86.91 \\ 
        dLoRA+FFT & dLoRA+FFT  & 37.54 & 17.63 & 15.39 & 83.64 \\ 
        \midrule
        \multicolumn{2}{c|}{\textit{Full model with pretraining}} & 42.56 & 18.24 & 18.02 & 90.11 \\
        \bottomrule
    \end{tabular}
    \vspace{-2mm}
\end{table}

\renewcommand{\arraystretch}{0.9}
\begin{table}[t]
\centering
\small
\caption{Question answering results with different LLMs on ScanQA~\cite{azuma2022scanqa}. \#Param/FLOP(G) denote the activated parameters and encoding/tokenization FLOPs. Best results in \textbf{bold}.
}
    \label{tab:sup_qa}
    \tabcolsep 1.4pt
    \vspace{-3mm}
    {%
    \begin{tabular}{lccc|cccc}
        \toprule
        \multirow{2}{*}{Method} & \multirow{2}{*}{Enc.} & \multirow{2}{*}{\#Params $\downarrow$} & \multirow{2}{*}{FLOP $\downarrow$} & \multicolumn{4}{c}{ScanQA (val)}\\
        \cmidrule(lr){5-8}
        & & & & R$\uparrow$ & M$\uparrow$ & B-4$\uparrow$ & C$\uparrow$\\
        \midrule
        \multicolumn{8}{c}{\textit{OPT-1.3B~\cite{zhang2022opt} LLM}} \\
        \midrule
        LL3DA~\cite{chen2024ll3da} & \cmark & 118.87M & 40.21 & 37.31 & 15.88 & 13.53 & 76.79\\
        PerLA~\cite{mei2025perla} & \cmark & 119.76M & 163.38 & 39.60 & 17.44 & 14.49 & 78.13\\
        \ourmethod & \xmark & \bf9.30M & \bf2.01 & \bf40.34 & \bf17.63 & \bf15.96  & \bf86.24\\
        \midrule
        \multicolumn{8}{c}{\textit{Qwen2.5-3B~\cite{qwen25} LLM}} \\
        \midrule
        LL3DA~\cite{chen2024ll3da} & \cmark & 118.87M & 40.21 & 37.24 & 16.01 & 14.91 & 79.18\\
        PerLA~\cite{mei2025perla} & \cmark & 119.76M & 163.38 & 39.91 & 16.08 & 15.53 & 81.42\\
        \ourmethod & \xmark & \bf10.54M & \bf2.04 & \bf42.56 & \bf18.24 & \bf17.12 & \bf90.11\\
        \bottomrule
    \end{tabular}
    }
    \vspace{-5mm}
\end{table}


\noindent\textbf{Different LLMs.}
We further evaluate \ourmethod\ with different language backbones, including OPT-1.3B~\cite{zhang2022opt} and Qwen2.5-3B~\cite{qwen25}. Here, \emph{Enc.} denotes variants using a pretrained 3D encoder, while \xmark\ indicates our encoder-free design that replaces the encoder with a lightweight MLP. 
As shown in \cref{tab:sup_qa}, \ourmethod\ matches or improves performance while drastically reducing 3D front-end cost. With OPT-1.3B on ScanQA, our encoder-free variant uses only 9.30M parameters and 2.01G FLOPs for encoding/tokenization, versus 118.87M/40.21G for LL3DA and 119.76M/163.38G for PerLA. Despite this reduction, it improves CIDEr from 78.13 to 86.24 (\gain{8.11}) and achieves the best ROUGE-L, METEOR, and BLEU-4. With Qwen2.5-3B, the trend remains: our model still uses only 10.54M parameters and 2.04G FLOPs, yet attains the best ScanQA validation performance (R 42.56, M 18.24, B-4 17.12, C 90.11). Overall, \ourmethod\ generalizes consistently across LLM backbones, while its encoder-free design matches or even outperforms encoder-based baselines at an order-of-magnitude lower computational cost.


\section{Conclusions}
We presented \ourmethod, an encoder-free, Fourier-based 3D LMM that addresses the twin challenges of scalability and permutation invariance for point clouds. By compactly representing scenes as structured superpoints, serializing them via SFCs, and applying an FFT-based context enhancer, our tokenizer efficiently approximates self-attention while preserving global context.
To further reduce the token count, we merge tokens using a sparse graph constructed through curve window-based voting.
In the language model head, Fourier-augmented LoRA injects frequency-aware interactions at negligible overhead, enabling strong performance without a heavy geometric backbone. Across experiments and ablations, \ourmethod matches or exceeds encoder-based 3D LMMs while substantially reducing computation and parameters.
\ourmethod inherits the limitations of serialization-based approaches such as PTv3~\cite{wu2024point}, which may underperform on non-Euclidean long-range relationships in highly cluttered scenes.
Future work includes pretraining on larger and more diverse 3D corpora, adaptive or learned serialization, and integration with additional modalities such as RGB images.

\subsection*{Acknowledgments}
This work was supported by the PNRR FAIR -- Future Artificial Intelligence Research project (PE00000013), funded by the European Union -- NextGenerationEU, and by the Ministero delle Imprese e del Made in Italy under the IPCEI Cloud initiative (DM 27 giugno 2022 -- IPCEI-CL-0000007). We also acknowledge ISCRA for granting access to the LEONARDO supercomputer, owned by the EuroHPC Joint Undertaking and hosted by CINECA (Italy).

{
    \small
    \bibliographystyle{ieeenat_fullname}
    \bibliography{main}
}

\clearpage
\setcounter{page}{1}
\maketitlesupplementary

\setcounter{section}{0}
\setcounter{table}{0}
\setcounter{figure}{0}

\renewcommand{\thesection}{\Alph{section}}
\renewcommand{\thetable}{\Alph{table}}
\renewcommand{\thefigure}{\Alph{figure}}

\section{Introduction}
In this supplementary material, we first present additional ablation studies and qualitative results that further demonstrate the effectiveness and efficiency of \ourmethod~(\cref{supp:analysis}). 
We then provide additional details of \ourmethod to complement the method description in the main paper~(\cref{supp:components}). 
Next, we elaborate on the computational complexity analysis of our proposed method (\cref{supp:efficiency}). 
Finally, we present the datasets used for training and evaluation, together with additional implementation details~(\cref{supp:dataset}).
We use the abbreviations C, B-4, M, and R to denote CIDEr~\cite{vedantam2015CIDEr}, BLEU-4~\cite{papineni2002bleu}, METEOR~\cite{banerjee2005meteor}, and ROUGE-L~\cite{lin2004rouge}, respectively.

\section{Additional results}
\label{supp:analysis}
\subsection{Number of LoRA layers}
We vary the number of LoRA-inserted layers (0, 4, 8, 10, 12), where 0 indicates that the LLM is not fine-tuned. The largest improvement is observed from 0 to 8 layers (C: \gain{12.9}, B-4: \gain{3.3}, M: \gain{2.9}, R: \gain{5.7}). With 10 layers, the QA accuracy reaches its peak (C = 87.05), while 8 layers achieves the best language metrics (B-4/M/R). Increasing to 12 layers leads to degraded performance, suggesting mild overfitting. Overall, 8--10 layers provide the best trade-off between QA accuracy and text quality. We use \#Layers = 8 in the final model to balance performance and computational cost. 
\begin{table}[h!]
\small
    \centering
    \caption{Ablation study on the number of LLM's LoRA layers.}
    \label{tab:ab_loss}
    \vspace{-3mm}
    \tabcolsep 12pt
    \resizebox{\columnwidth}{!}
    {%
    \begin{tabular}{ccccc} 
        \toprule
        \multirow{2}{*}{\#Layers} & \multicolumn{4}{c}{ScanQA (Validation)} \\
         & R$\uparrow$ & M$\uparrow$ & B-4$\uparrow$ & C$\uparrow$\\ 
        \midrule
        0  & 35.93 & 14.94 & 13.39 & 74.03 \\
        4  & 38.27 & 16.30 & 13.76 & 80.04 \\ 
        8  & 41.64 & 17.80 & 16.70 & 86.91 \\ 
        10 & 41.51 & 17.14 & 16.67 & 87.05 \\ 
        12 & 40.47 & 16.56 & 15.78 & 86.40 \\ 
        \bottomrule
    \end{tabular}
    }
    \vspace{-3mm}
\end{table}

\subsection{Space-filling curve count}
\begin{table}[t]
\centering
\small
\tabcolsep 12pt
\caption{Effect of the number of SFC curves $n_C$ in multi-curve serialization on ScanQA (val).}\label{tab_ab_sfc}
\vspace{-3mm}
\begin{tabular}{lcccc}
    \toprule
    $n_C$ & R$\uparrow$ & M$\uparrow$ & B-4$\uparrow$ & C$\uparrow$ \\
    \midrule
    1 & 37.42 & 15.96 & 14.23 & 79.06 \\
    2 & 40.49 & 16.78 & 15.96 & 85.64 \\
    4 & 41.64 & 17.80 & 16.70 & 86.91 \\
    6 & 41.72 & 17.93 & 17.01 & 87.14 \\
    \bottomrule
\end{tabular}
\end{table}
We analyze the effect of the number $n_C$ of space-filling curves (SFCs). As shown in \cref{tab_ab_sfc}, ScanQA performance improves as the number of SFCs increases, but largely saturates beyond $n_C=4$, with only marginal gains from $4$ to $6$. We therefore use $n_C=4$ as a good trade-off between accuracy and computational cost.

\subsection{Ablation studies on token merging}
\begin{table}[t]
\small
    \centering
    \caption{Ablation study on the number of LLM's input tokens.}
    \label{tab:ab_ntoken}
    \vspace{-3mm}
    \tabcolsep 12pt
    \resizebox{\columnwidth}{!}
    {%
    \begin{tabular}{ccccc} 
        \toprule
        \multirow{2}{*}{\#Token} & \multicolumn{4}{c}{ScanQA (Validation)} \\
         & R$\uparrow$ & M$\uparrow$ & B-4$\uparrow$ & C$\uparrow$ \\ 
        \midrule
        64  & 37.96 & 16.23 & 14.09 & 79.53 \\
        128 & 38.21 & 15.85 & 14.88 & 83.92 \\ 
        256 & 41.64 & 17.80 & 16.70 & 86.91 \\ 
        320 & 41.73 & 17.67 & 16.61 & 87.20 \\ 
        \bottomrule
    \end{tabular}
    }
\end{table}

We vary the number of LLM input vision tokens after merging (64, 128, 256, 320). Accuracy (C) improves with more tokens, with the largest gain up to 256 (\gain{7.38} C over 64; \gain{2.61} B-4; \gain{1.57} M; \gain{3.68} R). Beyond 256, gains saturate: 320 yields a small accuracy uptick (87.20 \vs.\ 86.91) and a negligible ROUGE change (41.73 \vs.\ 41.64), while BLEU-4 and METEOR slightly drop. Overall, 256 tokens provides the best language metrics (B-4/M) with near-peak accuracy, offering the best compute–quality trade-off. 
We set token number as 256 across our experiments.

\subsection{Ablation studies on two-stage training}
\cref{tab:sub_two_stage} reports the performance of \ourmethod\ under a two-stage setup compared with training from scratch. 
Two-stage training (generalist pre-training followed by ScanQA~\cite{azuma2022scanqa} instruction tuning) consistently improves all metrics: R increases from 41.64 to 43.37 (\gain{1.73}), M from 17.80 to 18.61 (\gain{0.81}), B from 16.70 to 16.87 (\gain{0.17}), and C from 86.91 to 91.74 (\gain{4.83}).
These gains indicate that a generic pre-training stage provides a stronger initialization for downstream ScanQA.
\begin{table}[t]
\small
    \centering
    \caption{Effect of two-stage training on ScanQA validation. 
    }
    \tabcolsep 12pt
    \vspace{-3mm}
    \begin{tabular}{ccccc}
        \toprule
        Method      & R$\uparrow$ & M$\uparrow$ & B-4$\uparrow$ & C$\uparrow$ \\
        \midrule
        Scratch     & 41.64 & 17.80 & 16.70 & 86.91 \\
        w/pretrain  & 43.37 & 18.61 & 16.87 & 91.74 \\
        \bottomrule
    \end{tabular}
    \label{tab:sub_two_stage}
\end{table}

\subsection{Comparison with token selection strategies}
To evaluate how different token selection and merging mechanisms affect both accuracy and token efficiency, 
we compare three strategies for reducing the number of 3D tokens:
(i) farthest point sampling (FPS),
(ii) a Q-Former-style cross-attention pooling module, and
(iii) our proposed graph based token merging.
All variants share the same backbone and are trained \emph{from scratch} on the ScanQA training set, 
\emph{without} any pre-training.
\begin{table}[t]
    \centering
    \small
    \caption{
    Effect of different token selection / merging strategies on ScanQA validation performance.
    All models are trained from scratch on the ScanQA training split, without any pre-training.
    }\label{tab:token_selector}
    \vspace{-3mm}
    \tabcolsep 10pt
    \begin{tabular}{lcccc}
        \toprule
        Selector   & R$\uparrow$ & M$\uparrow$ & B-4$\uparrow$ & C$\uparrow$ \\
        \midrule
        Q-Former   & 37.52 & 15.87 & 14.06 & 77.41 \\
        FPS        & 39.67 & 17.10 & 15.18 & 82.54 \\
        \ourmethod & \bf41.64 & \bf17.80 & \bf16.70 & \bf86.91 \\
        \bottomrule
    \end{tabular}
\end{table}
As shown in \cref{tab:token_selector}, FPS already improves over the Q-Former selector by 
about \gain{2.1} R, \gain{1.2} M, \gain{1.1} B-4, and \gain{5.1} C, 
indicating that purely geometric sampling yields stronger 3D language grounding than cross-attention pooling.
Our graph–based token merging further boosts performance across all metrics, 
achieving gains of roughly \gain{4.1} R, \gain{1.9} M, \gain{2.6} B-4, and \gain{9.5} C over the Q-Former baseline.

\subsection{More qualitative visualizations.}
\begin{figure}[t]
    \centering
    \includegraphics[width=0.48\linewidth]{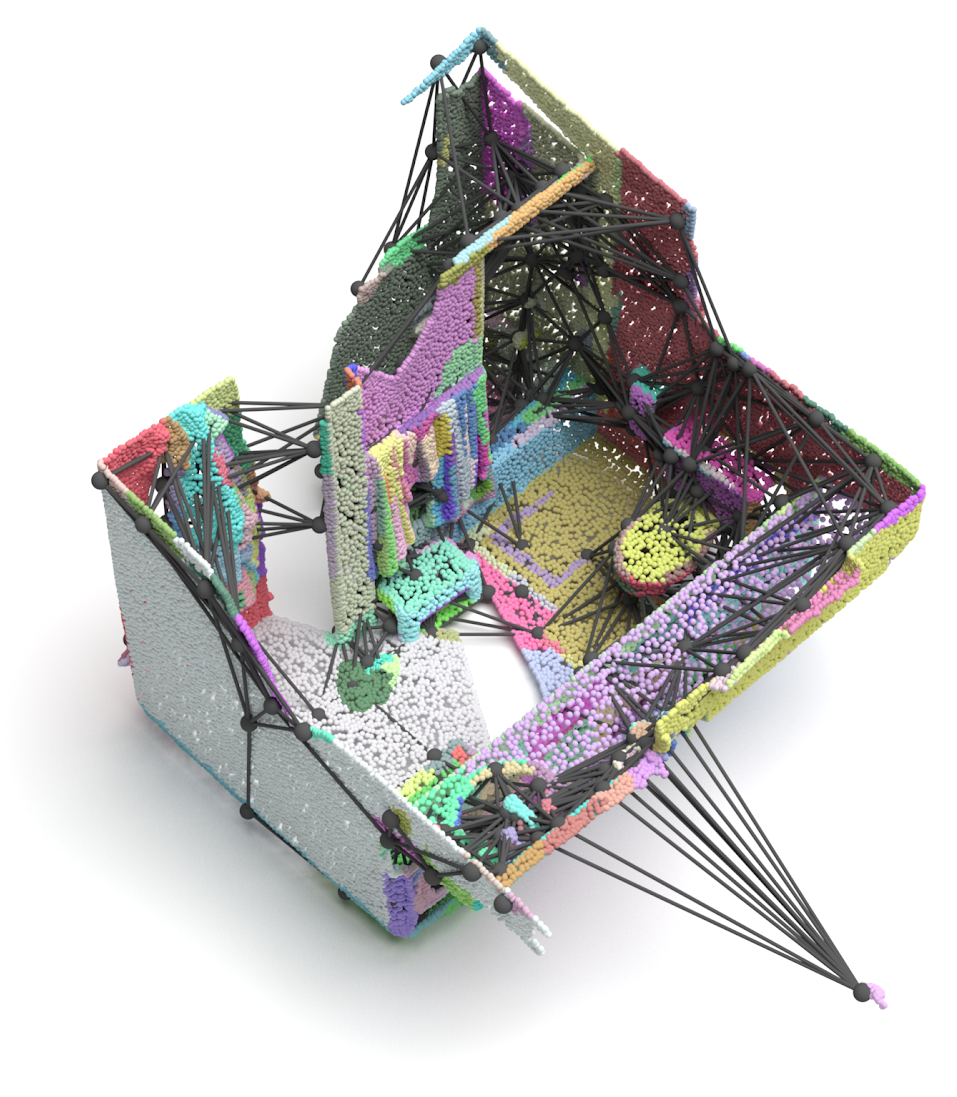}
    \hspace{1mm}
    \includegraphics[width=0.48\linewidth]{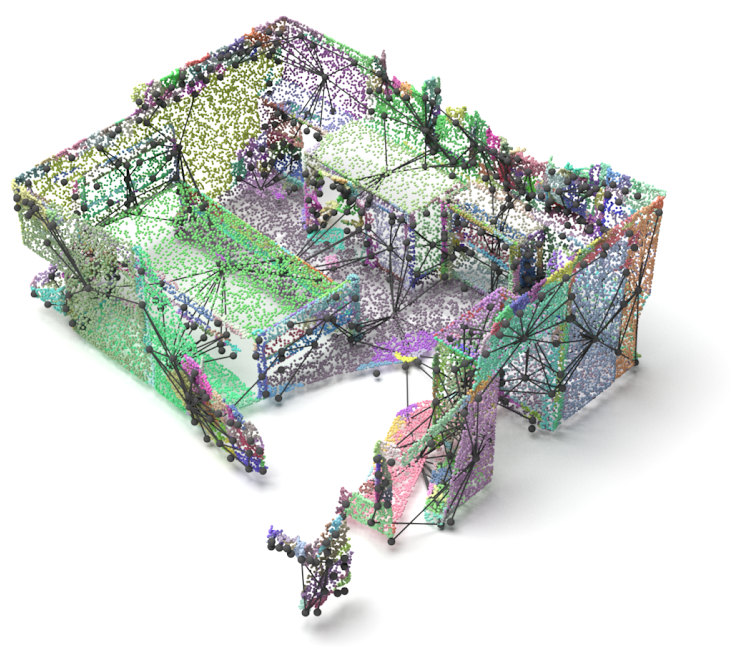}
    \vspace{-3mm}
    \caption{Visualization of our SFC-based $k$NN graph construction via window voting. 
    We show two representative examples of curve-guided neighbor selection.}
    \label{fig:sup_knn_searching}
    \vspace{-4mm}
\end{figure}

\cref{fig:sup_knn_searching} shows how we utilize space-filling curves (SFCs) to efficiently approximate $k$-NN search and construct our curve-based $k$-NN graph. The provided examples demonstrate that our method reliably identifies edges between superpoints, effectively preserving their genuine spatial connectivity.

\begin{figure*}[!hbt]
    \centering
    \tabcolsep 2pt
    \begin{tabular}{cccc}
        \begin{overpic}[width=0.24\linewidth]{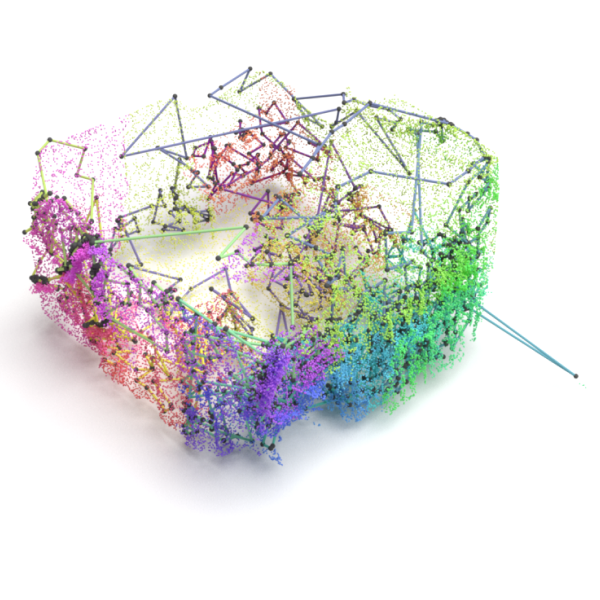}
            \put(-4,45){\rotatebox{90}{Hilbert}}
        \end{overpic} &
        \begin{overpic}[width=0.24\linewidth]{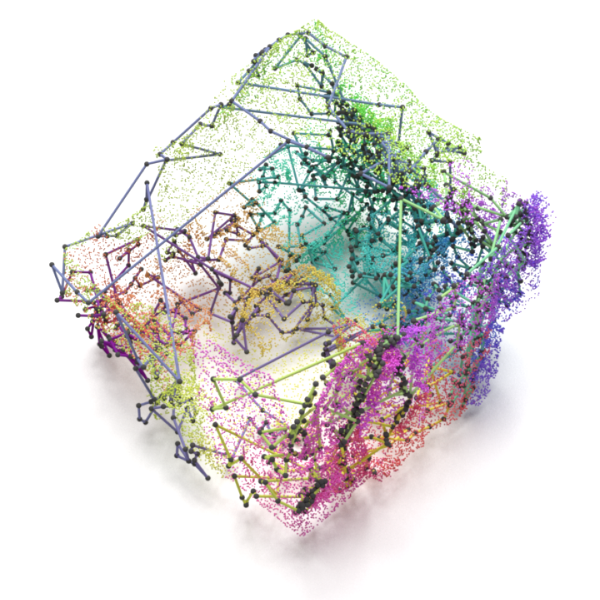}
        \end{overpic} &
        \begin{overpic}[width=0.24\linewidth]{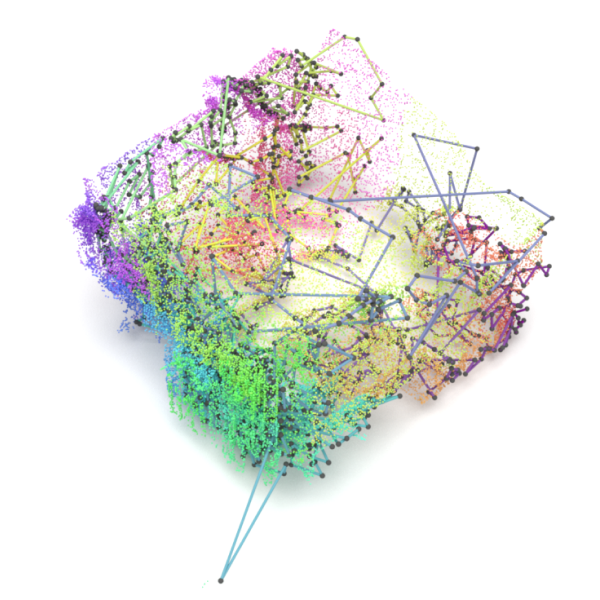}
        \end{overpic} &
        \begin{overpic}[width=0.24\linewidth]{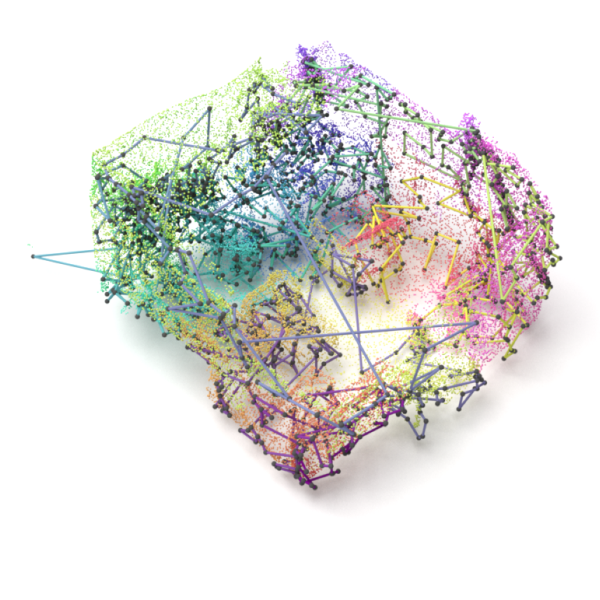}
        \end{overpic} \\[0.8ex]
        \begin{overpic}[width=0.24\linewidth]{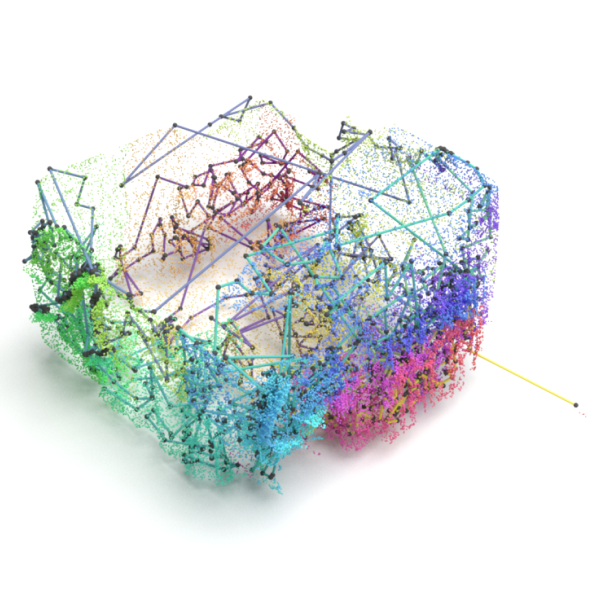}
            \put(-4,25){\rotatebox{90}{Transposed Hilbert}}
        \end{overpic} &
        \begin{overpic}[width=0.24\linewidth]{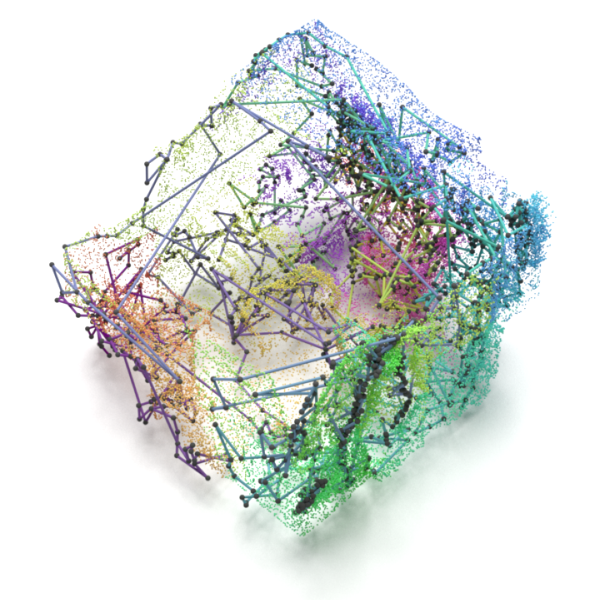}
        \end{overpic} &
        \begin{overpic}[width=0.24\linewidth]{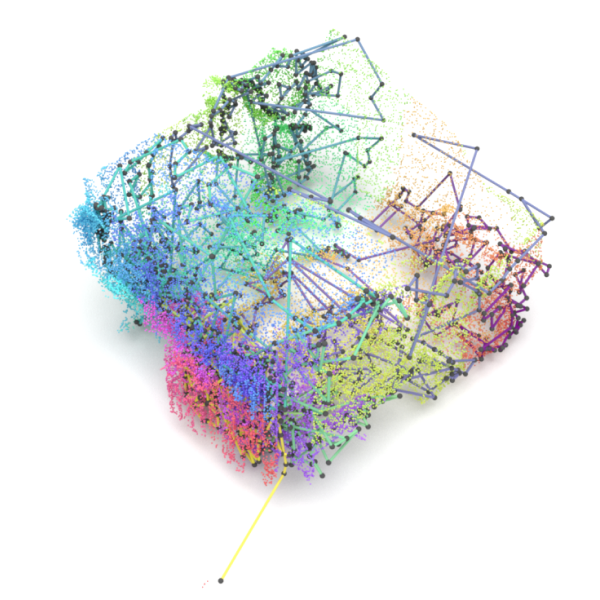}
        \end{overpic} &
        \begin{overpic}[width=0.24\linewidth]{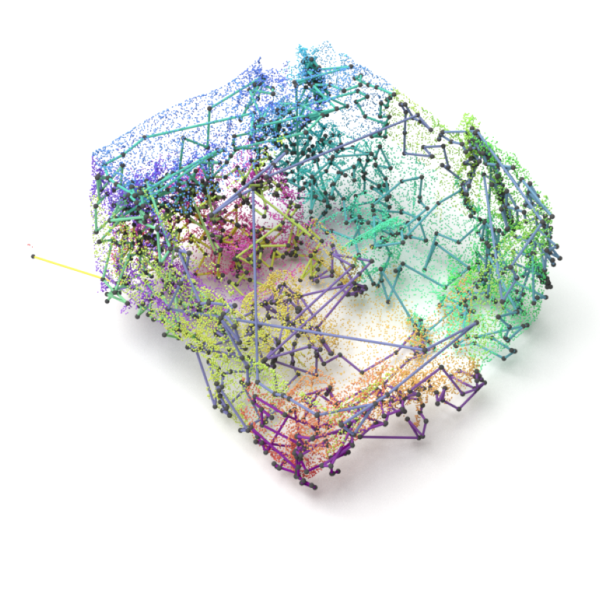}
        \end{overpic} \\[0.8ex]
        \begin{overpic}[width=0.24\linewidth]{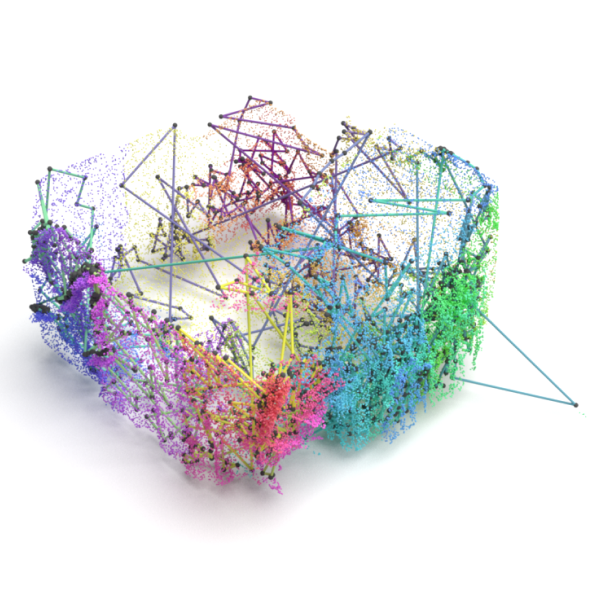}
            \put(-3,45){\rotatebox{90}{Z-order}}
        \end{overpic} &
        \begin{overpic}[width=0.24\linewidth]{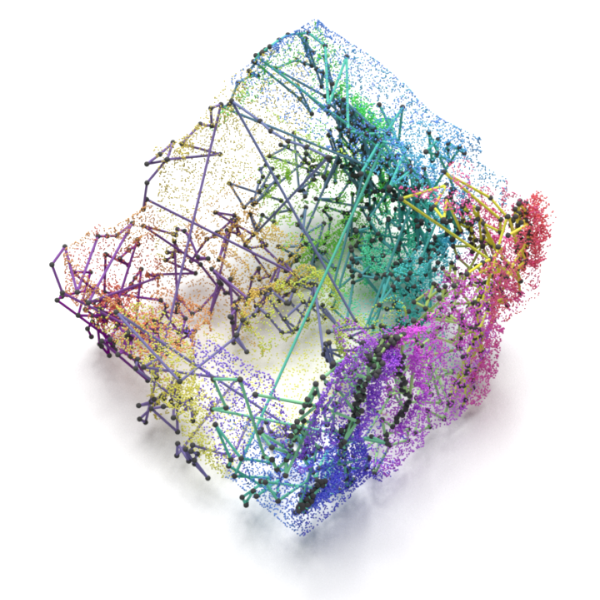}
        \end{overpic} &
        \begin{overpic}[width=0.24\linewidth]{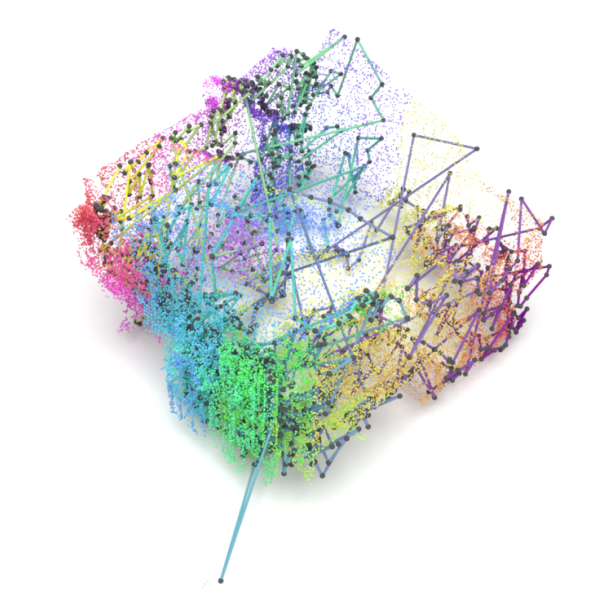}
        \end{overpic} &
        \begin{overpic}[width=0.24\linewidth]{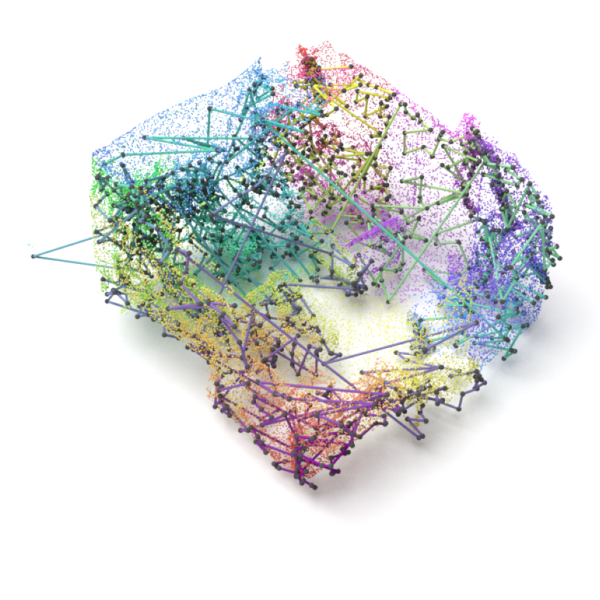}
        \end{overpic}\\[0.8ex]
        \begin{overpic}[width=0.24\linewidth]{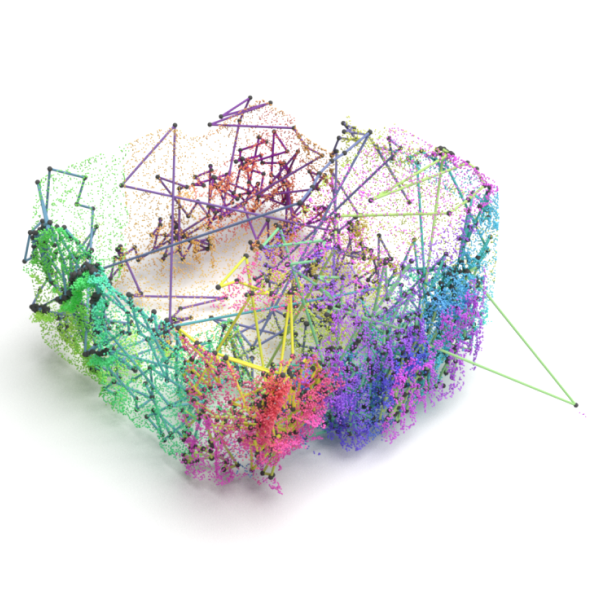}
            \put(-3,25){\rotatebox{90}{Transposed Z-order}}
        \end{overpic} &
        \begin{overpic}[width=0.24\linewidth]{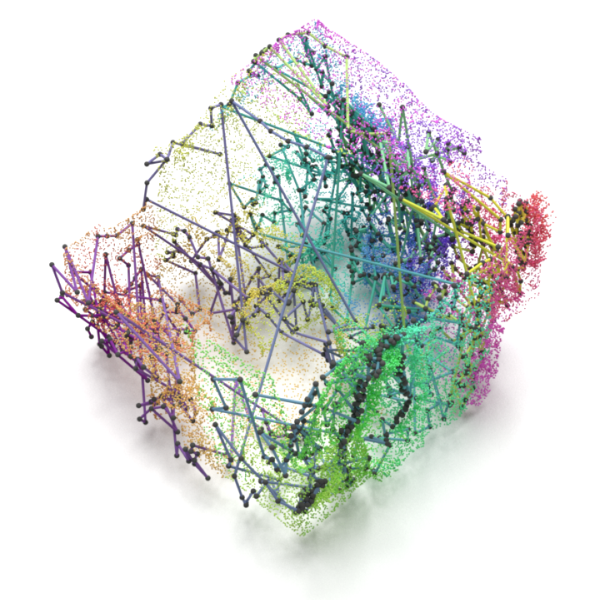}
        \end{overpic} &
        \begin{overpic}[width=0.24\linewidth]{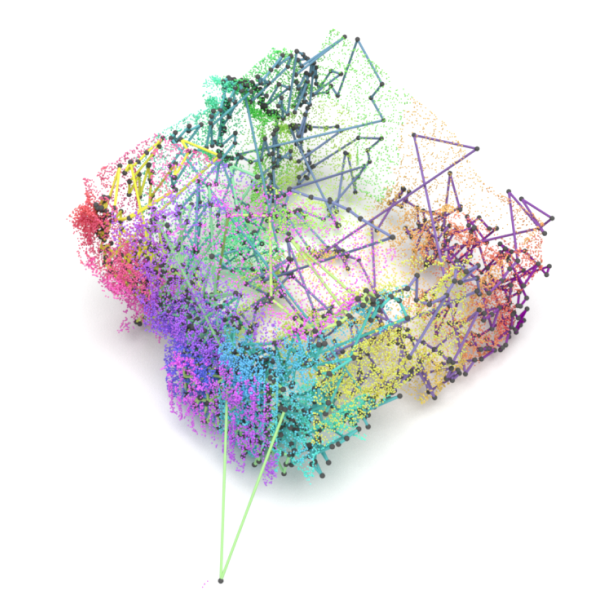}
        \end{overpic} &
        \begin{overpic}[width=0.24\linewidth]{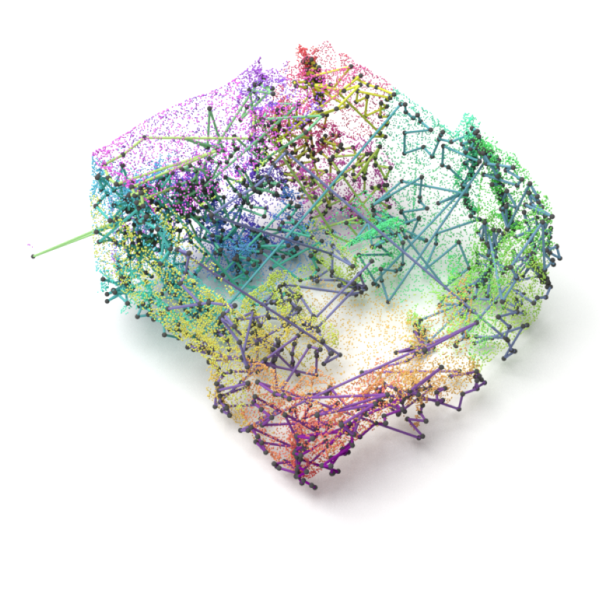}
        \end{overpic} 
    \end{tabular}
    \vspace{-4mm}
    \caption{Comparison of space-filling curves (Hilbert, transposed Hilbert, Z-order, and transposed Z-order) under four views (columns). All four variants produce broadly similar locality-preserving paths, with Hilbert-type curves appearing smoother and Z-order variants exhibiting more axis-aligned segments. Using both the standard and transposed versions provides complementary directional coverage and reduces orientation bias, so in \ourmethod we apply the FFT jointly over all four curves.}
    \label{fig:sup_sfc_comparison}
    \vspace{-3mm}
\end{figure*}

\cref{fig:sup_sfc_comparison} compares the space-filling curve (SFC) paths of four variants: Hilbert, transposed Hilbert, Z-order, and transposed Z-order. To make their differences easier to inspect, we render four viewpoints for each curve. While each variant traces the scene with a distinct trajectory, they all provide strong locality-preserving orderings. Specifically, Hilbert-based curves are smoother and exhibit fewer long-range jumps, whereas Z-order variants favor more axis-aligned runs. Including both the standard and transposed variants ensures complementary directional coverage, preventing orientation-specific artifacts and allowing the model to capture structural information consistently across different spatial alignments. This motivates our design choice in \ourmethod to use all four curves jointly within the FFT-based token enhancer, ensuring that these complementary locality patterns are seamlessly fused into a single spectral representation.

\section{Method details}
\label{supp:components}
\subsection{Fast Fourier transform}
We perform spectral mixing over the serialized token sequence using a real-valued frequency transform. 
In practice, the transform can be implemented with either FFT/iFFT-based operators or a DCT-style transform for improved efficiency on real-valued inputs.
Given a real-valued sequence $x = (x_n)_{n=0}^{N-1}$, its discrete Fourier transform (DFT) and inverse transform are written as~\cite{pfister2017discrete}
\begin{equation}
\begin{aligned}
\mathcal{F}(x)_k &= \sum_{n=0}^{N-1} x_n \exp\!\left(-i \frac{2\pi}{N} kn\right), \\
\mathcal{F}^{-1}(y)_n &= \frac{1}{N}\sum_{k=0}^{N-1} y_k \exp\!\left(i \frac{2\pi}{N} kn\right).
\end{aligned}
\end{equation}
For real-valued inputs, the transformed coefficients exhibit conjugate symmetry, \ie, $y_{N-k} = \overline{y_k}$. Therefore, the spectral transform can be implemented efficiently with reduced real-valued operators without changing the overall spectral mixing formulation.
To improve efficiency and reduce boundary artifacts caused by discontinuities in the serialized sequence, we process the sequence block-wise using an overlap--add scheme~\cite{pfister2017discrete}. 
Specifically, we partition the sequence into overlapping windows of length $L_w$ with stride $L_s$, reducing the complexity from $O(N\log N)$ to $O\!\left(N \log L_w\right)$. In our implementation, we use a Hann window with $L_s=L_w/2$ (\ie, 50\% overlap), which satisfies the constant overlap--add (COLA) condition and ensures stable reconstruction.
The smooth tapering at block boundaries also helps mitigate spectral leakage caused by abrupt jumps introduced by curve-based serialization.

\subsection{Token serialization}
Let $\mathcal{C} = \{c_i\}_{i=0}^{M-1}$ denote the 3D centers of the superpoints. We first normalize these centers relative to the scene bounding box and quantize them onto a $b$-bit integer grid:
\begin{equation}
\hat{c}_i 
= \big\lfloor (2^{b}-1)\,\frac{c_i - c_{\min}}{c_{\max}-c_{\min}} \big\rfloor,
\end{equation}
where $c_{\min}, c_{\max} \in \mathbb{R}^3$ are the minimum and maximum coordinates over all superpoint centers. 
Consequently, the quantized coordinates $\hat{c}_i = (\hat{c}_i^x,\hat{c}_i^y,\hat{c}_i^z)$ are bounded within the discrete space $\{0,\cdots,2^b-1\}^3$.
From these quantized coordinates, we compute a 1D key 
$k_i = \mathrm{SFC}(\hat{c}_i)$ 
using a space-filling curve (\eg, Z-order or Hilbert). This operation effectively maps the 3D spatial layout onto a 1D sequence, assigning each grid cell a unique integer index.

\subsection{More details about neighbor searching} 
Real-world captured point clouds often present holes, disconnections, or sparse regions. 
Consequently, the initial connectivity operation can yield spurious edges when attempting to merge object-level superpoints. 
To reduce incorrect neighbors, we employ a two-factor re-ranking strategy. 
We prioritize edges based on the \textit{shortest 3D Euclidean distance} as the primary factor. 
We then use the initial \textit{votes} (\ie, connectivity scores) as a secondary tie-breaker, retaining only the top $k$ nearest neighbors to construct the final, clean graph.
Specifically, we sum the neighborhood connectivity votes to obtain a coalesced candidate edge set $\mathcal{E}^{\text{cand}}=\{(s,t,v_{s,t})\,|\,v_{s_t}>0\}$. 
To suppress spurious connections resulting from SFC topological folding, we employ a re-ranking strategy for each source node $s$. 
Candidates are ranked using the squared Euclidean center distance $d_{s,t}=\|c_s-c_t\|_2^2$ as the primary sorting key, and the connectivity votes $v_{s,t}$ as the secondary tie-breaker. 
We then retain the $k$ nearest neighbors that minimize this composite tuple $(d_{s,t}, -v_{s,t})$. 
The resulting sparse graph $\mathcal{G}$, represented by its adjacency matrix $A$ and degree matrix $D$.

\section{Computational complexity analysis}\label{supp:efficiency}
\subsection{Graph construction}
Superpoint neighbors in our setting are defined by the minimum distance between their \emph{boundary points}, \ie, points lying near inter-superpoint interfaces. Let $N$ be the number of points in a scene, $M$ the number of superpoints, and $B$ the batch size. Constructing a superpoint adjacency graph is a critical step in 3D tokenization, as it largely determines both the computational cost and the memory footprint of the overall vision--language pipeline.

In this section, we analyze and compare the computational complexity of our proposed \emph{multi-curve voting graph} against a conventional \emph{raw superpoint $k$-nearest-neighbor ($k$-NN)} graph constructed from boundary-based distances.

\paragraph{Raw KNN-based graph on superpoints.}
A brute-force construction compares points across all \emph{distinct} superpoints to identify potential boundary interactions. Boundary-aware distances are computed \emph{a posteriori} by aggregating inter-superpoint point-to-point distances (\eg, taking the minimum), followed by a top-$k$ selection step.
Specifically, let $p_i$ denote the number of points in superpoint $i$, such that the total number of points is $N = \sum_i p_i$. 
A naive implementation scanning all point pairs across distinct superpoints incurs a computational cost of
\begin{equation}\label{eq:sup_knn}
    T_{\text{raw}}
    = O\Big(
        B \sum_{i \ne j} p_i p_j
      \Big).
\end{equation}
Applying the algebraic identity $\sum_{i \neq j} a_i a_j =(\sum_i a_i)^2 - \sum_i a_i^2$, \cref{eq:sup_knn} then simplifies to:
\begin{equation}
    T_{\text{raw}} 
    = O\Big( B \big(N^2 - \textstyle\sum_i p_i^2\big) \Big).
\end{equation}
In any typical over-segmentation ($M \geq 2, N_M \leq p_i \leq N / 2$), the points are distributed across multiple regions, ensuring that the intra-superpoint term is negligible compared to the total interactions ($\sum_i p_i^2 \ll N^2$). $N_M\geq 2$ is a predefined threshold.
Consequently, $T_{\text{raw}} = O(B N^2) \gg O(N \log N)$, this brute-force approach asymptotically dominates any efficient near-linear spatial indexing scheme.
The memory footprint is likewise prohibitive, requiring the materialization of a dense $M \times M$ distance structure as $M_{\text{raw}} = O\big(B M^2\big).$
Even for moderate over-segmentation (\eg, $M{=}1024$), this yields over $10^6$ entries per scene. 
Furthermore, because boundary points are not known \emph{a priori}, the boundary-aware variant amplifies load imbalance: superpoints with complex interfaces or higher densities disproportionately increase the constant factors in the runtime.

\paragraph{Multi-curve sparse voting graph.}
Our method sidesteps dense global pairwise comparisons by projecting the point cloud onto multiple space-filling curves and restricting neighbor search to short 1D windows along each curve.
Let $C {=} |\pi|$ be the number of curves, $W$ the local window radius on each curve (so each window has at most $2W{+}1$ points), and $s_r$ the point-sampling stride along the curves.
Each source point participates in at most $(2W{+}1)$ 1D neighbors per curve, and we only evaluate these local pairs.
Across all curves and all batches, the total number of candidate edges is
\begin{equation}
    E \;\approx\;
    C \cdot B \cdot \frac{N}{s} \cdot (2W{+}1)
    \;=\;
    O\Big(
      C B \frac{N W}{s}
    \Big).
\end{equation}
Consequently, the number of candidate edges grows \emph{linearly} with $N$, rather than quadratically. We then aggregate all per-curve votes into a sparse coordinate (COO) tensor and resolve duplicates via a multi-key stable sort. This yields an overall time complexity of:
\begin{equation}
    T_{\text{curve}}
    =
    O\Big(
      C B \frac{N W}{s}
      \log\big( C B \tfrac{N W}{s} \big)
    \Big),
\end{equation}
which is near-linear with respect to the total number of points, bounded only by a small logarithmic sorting factor. The subsequent per-superpoint top-$k$ selection operates strictly on the outgoing edges of each node. 
This introduces a lower-order term of $O(B M k \log k)$, which is asymptotically negligible compared to the global sparse sorting over $E$. 
The memory usage scales linearly with the scene size:
\begin{equation}
    {M}_{\text{curve}}
    =
    O(C B N) + O(E).
\end{equation}
Since all intermediate structures remain sparse, we avoid instantiating a dense $M \times M$ adjacency matrix. 

Overall, our multi-curve voting graph replaces the quadratic time and memory complexity of a naive $k$-NN graph construction with a highly efficient, near-linear alternative. This design ensures robust scalability to high-resolution scenes with a large number of superpoints.

\subsection{FFT-based context enhancer}
Let $D$ be the feature dimension, and $N_h$ the number of frequency heads (with per-head width $d_h \approx D/N_h$).
We analyze the computational complexity of the FFT-based context enhancer along the superpoint axis, specifically focusing on the windowed variant utilized in our implementation.

\paragraph{Windowed FFT mixing.}
Given a sequence of $M$ tokens, the mixer partitions the input into overlapping windows of length $L_w$ and stride $L_s$. The number of windows is
\begin{equation}
    N_w =\Big\lfloor \frac{M - L_w}{L_s} \Big\rfloor + 1
    \;=\; O\!\left(\frac{M}{L_s}\right).
\end{equation}
For each window, we apply a 1D FFT of length $L_w$ on every head, perform frequency-domain gating, and then apply the inverse FFT. Since each head has dimension $d_h$ and there are $N_h$ heads with $N_h d_h {=} D$, the total cost of the operations is
\begin{equation}
    T_{\text{FFT}}
    =
    O\!\left(
        B \,\frac{M}{L_s}\, D \, L_w \log L_w
    \right).
\end{equation}
The remaining steps, including window weighting, linear input/output projections, and overlap-add reconstruction, contribute only linear overhead.
When $L_w$ and $L_s$ are fixed constants, as in our experiments, the dominant cost simplifies to $T_{\text{FFT}} = O(BMD),$
which is effectively linear in the number of tokens $M$. The memory complexity is also linear as $\mathcal{M}_{\text{FFT}} = O(BMD),$
which accounts for the unfolded windows, FFT buffers, and overlap-add reconstruction buffers.

\paragraph{Multi-curve fusion.}
When operating on multiple space-filling curves (\eg, Hilbert, transposed Hilbert, Z-order, transposed Z-order) with $|\pi|$ curve traversals, we apply the same windowed FFT mixer independently to each curve and fuse their outputs.
This introduces a multiplicative factor in both time and memory:
\begin{equation}
T_{\text{FFT}}
    = O\big(|\pi| \, B D M\big), 
    M_{\text{FFT}}
    = O\big(|\pi| \, B M D\big).
\end{equation}
The subsequent curve-attention scorer introduces only a lightweight $O(B|\pi|D)$ overhead, which remains asymptotically negligible compared to the core FFT operations.

\section{Dataset and Implementation details}\label{supp:dataset}
\subsection{Dataset}
During training, we use the ScanNet portion of the 3D-LLM dataset~\cite{hong20233dllm} as our primary source. 
We further augment it with complementary datasets, including ScanQA~\cite{azuma2022scanqa}, SQA3D~\cite{ma2022sqa3d}, ScanRefer~\cite{chen2020scanrefer}, and Nr3D~\cite{achlioptas2020referit3d}. 
After joint training on this mixture, we fine-tune \ourmethod\ separately on each target dataset. 
We briefly summarize the datasets below.

\paragraph{3D-LLM dataset~\cite{hong20233dllm}} 
comprises: (i) 1,033 textual descriptions across 517 scenes, 
(ii) 1,864 lines of embodied task planning spanning 510 scenes, and 
(iii) 2,955 lines of multi-turn embodied dialogues across 517 scenes.

\paragraph{ScanQA dataset~\cite{azuma2022scanqa}} 
is a 3D question answering benchmark built on top of ScanNet~\cite{dai2017scannet}, designed to evaluate scene understanding via natural language queries. 
It contains 6,857 unique questions paired with 30,769 answers over 806 reconstructed indoor environments. 
Questions focus on objects in the scene and cover object attributes, spatial relationships, and scene semantics. 
On average, each scene has 8.5 questions, encouraging reasoning over object-level details and contextual relations in complex 3D environments.

\paragraph{ScanRefer dataset~\cite{chen2020scanrefer}} 
is a multimodal benchmark for 3D vision-language reasoning built on ScanNet~\cite{dai2017scannet}, consisting of 1,613 RGB-D scans across 806 unique indoor scenes. 
It provides 51,583 natural language descriptions for objects in reconstructed 3D scenes, covering 800 ScanNet scenes. 
Each object is annotated with an average of 4.67 descriptions, resulting in rich linguistic diversity. 
On average, each scene contains 13.81 objects and 64.48 descriptions, spanning over 250 categories of common indoor objects. 
Among these, 41,034 descriptions explicitly mention attributes such as color, shape, size, and spatial relations, making ScanRefer a strong benchmark for fine-grained language reasoning.

\paragraph{Nr3D dataset~\cite{achlioptas2020referit3d}} 
is a natural language 3D object localization benchmark built on ScanNet~\cite{dai2017scannet}. 
It contains 41,503 unique referring expressions for 5,578 objects across 707 scenes. 
Each description is crafted to unambiguously identify a target object in context, using spatial relations and attributes such as color, shape, and size. 
On average, each object is associated with 7.4 descriptions, making Nr3D well suited for evaluating fine-grained attribute understanding and spatial reasoning in 3D scenes.

\paragraph{SQA3D dataset~\cite{ma2022sqa3d}} 
is a spatial question answering benchmark in 3D environments, built upon ScanNet~\cite{dai2017scannet}. 
It comprises natural language questions that require explicit 3D spatial reasoning, including object existence, relative positions, and relationships among multiple objects. 
Each question is grounded in a reconstructed 3D scene and paired with a free-form textual answer, enabling the evaluation of both geometric understanding and language generation. 
Compared to traditional 2D VQA benchmarks, SQA3D places greater emphasis on 3D-aware reasoning over scene layouts, object configurations, and embodied scene understanding, making it a challenging testbed for holistic 3D VLMs.

\subsection{Implementation details}
\paragraph{Superpoint generation.} We directly use the segmentor provided by ScanNet~\cite{dai1901transformer}, which is obtained via a graph-cut–based over-segmentation method~\cite{hui2022graphcut,felzenszwalb2004efficient}.

\paragraph{Training strategy.}
Following prior work~\cite{chen2024ll3da,mei2025perla}, \ourmethod adopts a two-stage training strategy. 
In the first stage, we pre-train the model on an ensemble of datasets spanning diverse 3D tasks (dense captioning and question answering), so that it acquires a broad understanding of 3D scenes and functions as a 3D \emph{generalist} model. 
In the second stage, we perform instruction-following fine-tuning on task-specific datasets for 3D dense captioning and 3D question answering, thereby specializing the model for the downstream benchmarks.
Note that during both stages we do not use proposals from Mask3D~\cite{schult2023mask3d}; all supervision is applied directly on our own tokenization pipeline.

\paragraph{Hyperparameters.}
Unless otherwise specified, for graph construction, window-based voting uses a window size of 64 and a stride of $s_r=16$ along each space-filling curve. The number of initial superpoints, $M$, depends on the input point cloud and thus varies across samples. For the prompt embedding, we set 16 neighbors for the $k$-NN searching.

\end{document}